\documentclass{article}

\usepackage[margin=1.3in]{geometry}
\usepackage{amsmath,amsthm,amssymb}

\usepackage{parskip}
\usepackage{subcaption}
\usepackage{graphicx}
\usepackage{xspace}
\usepackage{authblk}
\usepackage{physics}
\usepackage{algpseudocode}
\usepackage{todonotes}

\newtheorem{lemma}{Lemma}
\newtheorem{proposition}{Proposition}

\theoremstyle{definition}
\newtheorem{assumption}{Assumption}

\newcommand{\modelprop}{\textsc{ModelProportionalSampling}\xspace} 
\newcommand{\mps}{\textsc{MPS}\xspace}
\newcommand{\ent}{\textsc{Ent}\xspace}
\newcommand{\kl}{\textsc{KL}\xspace}
\newcommand{\srs}{\textsc{SRS}\xspace}

\newcommand{\A}{\mathcal{A}}

\newcommand{\X}{\mathcal{X}}
\newcommand{\D}{\mathcal{D}}
\newcommand{\R}{\mathbb{R}}
\renewcommand{\Pr}{\mathbb{P}}
\newcommand{\Var}{\mathbb{V}}

\newcommand{\E}{\mathbb{E}}

\newcommand{\bpi}{\boldsymbol\pi}

\newcommand{\ind}{\textbf{1}}
\renewcommand{\S}{\mathcal{S}}

\newcommand{\Cov}{\text{Cov}}

\newcommand{\model}{\hat{\varphi}}
\newcommand{\re}{r}

\newcommand{\reward}{\textsc{Rew}}
\newcommand{\softmax}{\textnormal{softmax}}
\newcommand{\kld}[2]{D_{\textsc{kl}}(#1||#2)}
\newcommand{\hpi}{\hat{\pi}}

\newcommand{\modelmin}{\model_{\textnormal{min}}}

\newcommand{\tot}{\textsc{Pop}}
\newcommand{\htot}{\widehat{\tot}}
\newcommand{\htotmod}{\htot_{\textsc{model}}}

\newcommand{\htotipw}{\htot_{\textsc{ipw}}}
\newcommand{\htotdr}{\htot_{\textsc{dr}}}

\usepackage[square]{natbib}
\usepackage{algorithm}
\usepackage[export]{adjust box}

\usepackage[pagebackref=true]{hyperref}
\hypersetup{
colorlinks=true,
linkcolor=blue,
citecolor=blue
}
\setcitestyle{numbers}

\title{Entropy Regularization for Population Estimation}
\author[1,2]{Ben Chugg}
\author[2]{Peter Henderson}
\author[3]{Jacob Goldin}
\author[2]{Daniel E. Ho}
\affil[1]{Carnegie Mellon University}
\affil[2]{Stanford University}
\affil[3]{University of Chicago}
\date{}

\begin{document}
\maketitle

\begin{abstract}
Entropy regularization is known to improve exploration 
in sequential decision-making problems.
We show that this same mechanism can also lead to nearly unbiased and lower-variance estimates of the mean reward in the optimize-and-estimate structured bandit setting.
Mean reward estimation (i.e., population estimation) tasks have recently been shown to be essential for public policy settings where legal constraints often require precise estimates of population metrics.
We show that leveraging entropy and KL divergence can yield a better trade-off between reward and estimator variance than existing baselines, all while remaining nearly unbiased. 
These properties of entropy regularization illustrate an exciting potential for bridging the optimal exploration and estimation literatures. 

\end{abstract}

\section{Introduction}
\label{sec:introduction} 

While most frameworks for online sequential decision-making focus on the objective of maximizing reward, in practice this is rarely the sole objective. Other considerations may involve budget constraints, ensuring fair treatment, or estimating various population characteristics. There has been growing recognition that these other objectives must be formally integrated into sequential decision-making frameworks, especially if such algorithms are to be used in sensitive application areas~\citep{henderson2021beyond}. In this work, we focus on the problem of maximizing reward while simultaneously estimating the population total (equivalently, mean) in a structured bandit setting.

The most natural approach to this problem from a machine learning perspective is to use a model to predict the mean. However, this method is subject to the problem that adaptively collected data are subject to bias, which in turn biases the model estimates~\citep{nie2018adaptively}. 
Natural tools from survey sampling such as IPW estimators (also used commonly in off-policy evaluation (OPE)) enable mean estimation if
each observation has been sampled with some known or estimable probability. However, in those settings the probabilities are given \emph{a priori},
and can yield high-variance estimates if the sampling distribution is skewed.
Here we seek to optimize over possible probability distributions in order to trade off between our expected reward and the variance of our estimator.
Unfortunately, the variance formulation of such estimators is unwieldy and yields an intractable optimization problem if directly incorporated into the objective. Instead, we substitute the variance term for an information-theoretic policy regularization term such as entropy or KL divergence. Adding such terms yields analytically tractable optimization problems, while maintaining our ability to smoothly navigate reward-variance trade offs.

In reinforcement learning (RL) settings, adding policy regularization terms to the objective function has been shown to consistently improve the performance of sequential decision-making algorithms~\citep{williams1991function,ahmed2019understanding,xiao2019maximum}. Our ability to leverage these tools in a different setting enables us to take advantage of well-established theory and insights in RL, indicating an exciting convergence between our problem, survey sampling, and other areas of sequential decision-making.

In sum, our contributions are as follows: (i) We propose novel algorithms to handle the dual objective of reward maximization and mean reward estimation in the structured bandit setting; (ii)  characterize the bias of the estimators in our setting, provide closed-form solutions for our proposed optimization problems, and relate entropy and KL sampling strategies to the variance of the estimators; and  (iii) demonstrate the improvement of our algorithms over baselines on four datasets. Replication code and data can be found at \url{https://github.com/bchugg/ent-reg-pop-est}. 

\subsection{Related Work}

The bandit literature is large. We focus on the most relevant related works here, and relegate a prolonged discussion of all peripheral work to Appendix~\ref{app:related_work}. 

The closest work to ours is  \citet{henderson2022integrating}, who introduce the \emph{optimize-and-estimate} structured bandit setting which we adopt in this paper. They introduce Adaptive Bin Sampling (ABS) which we describe below and use as our main baseline. The optimize-and-estimate setting is closest to that in \citet{abbasi2011improved} and \citet{joseph2016fair}, but extended to  non-linear rewards with a required estimation objective. Crucially, in this setting, arms may change from step to step and the agent must instead rely on a per-arm context to determine which arms to pull.

A number of works have sought to yield reduced-bias (e.g., \citet{nie2018adaptively}) or unbiased estimators (e.g., \citet{zhang2020inference}) for inference with adaptively-collected data.
This is because it is well known that adaptive sampling leads to bias when estimating sample means and other population characteristics~\citep{nie2018adaptively,shin2019sample,shin2020conditional,shin2021bias,russo2016controlling}. Attempts to remedy this bias have come in the form of differential privacy~\citep{neel2018mitigating}, adaptive estimators~\citep{dimakopoulou2021online,chugg2021reconciling}, or re-normalization~\citep{zhang2020inference}.
Unfortunately, none of these efforts apply to our setting. They are mostly in the multi-armed and contextual bandit setting (wherein one samples the same arm multiple times) which differs from our own structured bandit setting. 
We emphasize that we are in the non-linear setting and thus seek to have non-parametric unbiasedness guarantees. Moreover, much previous work seeks to remedy bias \emph{ex post}, as opposed to incorporating bias (and variance) into the objective \emph{ex ante.}

Finally, a large body of work has examined entropy regularization for optimal exploration in reinforcement learning~\citep{williams1991function, zimmert2021tsallis,ahmed2019understanding,brekelmans2022your} and bandits~\citep{fontaine2019regularized,xiao2019maximum}. While the problem settings are different, we examine the fundamental question of whether entropy regularization may bring the same benefits for unbiased estimation as it has brought for exploration guarantees elsewhere.

\section{Problem Setting}

\subsection{Optimize-and-estimate structured bandits}

At time $t$ we receive a set of $N_t$ observations with features $X_t=\{x_{it}\}\subset\X$, where observation (``arm'') $x$ has reward $\re(x)\sim\D(x)$ drawn from a reward distribution conditional on the context. We assume that $r(x)\in[0,1]$ for all $x$. Each period $t$ (we use period and time step interchangeably) we can sample at most $K_t$ observations and receive their rewards. If $K_t=1$ we are in the \emph{sequential} setting; if $K_t>1$ we are in the \emph{batched} setting. Let $S_t\subset X_t$, $|S_t|\leq K_t$ denote the set (possibly singleton) of observations selected. Like in traditional bandit problems, one of our objectives is to maximize reward (equivalently, to minimize regret). The cumulative reward at time $T$ is 
\begin{equation}
    \reward(T) = \sum_{t\le T}\sum_{x\in X_t}\re(x)\ind_{x\in S_t},
\end{equation}
where $S\subset X $ is the set of selected observations. For an event $E$, the function $\ind_E$ is 1 if $E$ occurs and 0 otherwise.  Our second goal is to achieve a reliable population estimate $\tot(T)$ of the total of all arms in each period $T$:
\begin{equation}
    \tot(T) = \sum_{x\in X_T} \E_\D[\re(x)].
\end{equation}
Note that previous work in this setting estimated the population \emph{mean}, instead of the total. Obviously, the difference is unimportant, and we find it cleaner to work with the total. 

Throughout, we will drop the parenthetical in favor of a subscript when space demands and write, e.g., $r_x$ in lieu of $r(x)$. If the timestep $T$ is clear from context we drop it from the notation. 

We observe that observations are equivalent to arms and can be volatile from timestep to timestep, characterized only by a context and an underlying shared reward structure. This is similar to the structured bandit~\citep{abbasi2011improved,joseph2016fair}. If arms were fixed and the context singular, this would reduce to the contextual bandit. If the context were removed and arms remained fixed, this would reduce to the multi-armed bandit. As such, methods from the multi-armed bandit and contextual bandit (like Thompson sampling) do not necessarily apply to the structured bandit, and especially not the optimize-and-estimate structured bandit.

\subsection{Strategies for Population Estimation}

Given a model $\model:\X\to\R$ which estimates the reward $r(x)$, a natural approach to estimate $\tot$ is to combine the empirical population total of the selected points with the estimated total from $\model$: 
\begin{equation}
    \label{eq:model_est}
    \htotmod(T) = \sum_{x\in S_T} \re(x) + \sum_{x\in X_T\setminus S_T} \model(x). 
\end{equation}

A separate approach is available if selection is performed according to a probability distribution 
$\{\pi(x)=\Pr(x\in S)\}$ over the observations. In this case we can turn to ``importance sampling methods'', which employ the basic idea of weighting the observations by a function of their probability. We'll focus on two popular importance sampling methods: inverse propensity weighting (IPW) and doubly-robust (DR) estimation. 

Suppose that $x$ was sampled with probability $\pi(x)$ and that we have estimates $\hpi(x)$ of $\pi(x)$ for all $x$ (we allow for the possibility that $\hpi(x)=\pi(x)$, but we'll see the utility of allowing approximations later on). Assuming that $\hpi(x)>0$ for all $x$, then the IPW estimator~\citep{horvitz1952generalization,narain1951sampling} is 
\begin{equation}
    \label{eq:ipw}
    \htotipw(T) = \sum_{x\in X_T}\frac{
    \re(x)}{\hpi(x)}\ind_{x\in S_T}. 
\end{equation}
The doubly-robust (DR) estimator~\citep{cassel1976some,jiang2016doubly}, on the other hand, combines the model-based approach with the IPW estimator: 
\begin{equation}
    \label{eq:dr}
    \htotdr(T) = \sum_{x\in X_T}\bigg(\model(x) + \frac{\re(x) - \model(x)}{\hpi(x)}\ind_{x\in S_T}\bigg). 
\end{equation}
The IPW and DR estimators are common in off-policy evaluation~\citep{dudik2011doubly,dudik2014doubly}. The bias of all three estimators is given in Lemma~\ref{lem:bias}. While the bias was previously given by \citet{dudik2014doubly}, that was in the RL setting, which differs from ours. That said, the results do not change much. 

\begin{lemma}[Bias of estimators]
\label{lem:bias}
Let $\Delta_x=\E_\D[r(x)]-\model(x)$, and $\lambda_x = \pi(x)/\hpi(x)$. Then, at any time $T$, 
\begin{enumerate}
    \item $|\E_{\D,S}[\htotmod] - \tot| = |\sum_{x\in X_T}\Delta_x(\pi_x-1)|$. 
    \item $|\E_{\D,S}[\htotipw] - \tot|= |\sum_{x\in X_T}\E_\D[r_x](\lambda_x-1)|$, and 
    \item $|\E_{\D,S}[\htotdr] - \tot| = |\sum_{x\in X_T}\Delta_x(\lambda_x-1)|$.
\end{enumerate}
\end{lemma}
The proof of Lemma~\ref{lem:bias}, along with all other propositions in the paper, can be found in the Appendix.
Note a corollary of Lemma~\ref{lem:bias}: IPW and DR are unbiased if $\hpi(x)=\pi(x)$, i.e., we sample with precise inclusion probabilities. This is computationally challenging for sufficiently large budgets, but as we discuss in Section~\ref{sec:pareto}, we employ an approximation mechanism -- Pareto Sampling -- that, in practice, enables unbiasedness (Figure~\ref{fig:bias_ipw}).

\subsection{Variance}

The probabilities $\pi(x)=\Pr(x\in S)$ are called (first order) \emph{inclusion probabilities}. 
The terms $\pi(x,z) = \pi_{x,z} =  \Pr(x,z\in S)$ are called \emph{second order}, or \emph{joint} inclusion probabilities, and naturally arise in the variance of population estimators due to the covariance terms. 
More detail on inclusion probabilities can be found in Appendix~\ref{app:incl_probs}. 

To define the variance, for an arbitrary function $\theta:\X\to\R$, let
\begin{equation}
\label{eq:general_variance}
 A_T(\theta) = \frac{1}{2}\sum_{x,z\in X_T}\bigg(\frac{\theta(x)}{\pi(x)} - \frac{\theta(z)}{\pi(z)}\bigg)^2(\pi(x)\pi(z)-\pi(x,z)).   
\end{equation}
The variance of $\htotipw$ and $\htotdr$ (with respect to sampling) for $\hpi(x)=\pi(x)$ can then be written as (see Appendix~\ref{app:variance} for a derivation)
\begin{align}
\label{eq:variances}
    \Var_\pi(\htotipw(T)) = A_T(\re), \\ 
    \Var_\pi(\htotdr(T)) = A_T(\re - \model).
\end{align}
That is, fixing the inclusion probabilities, the variance of the IPW estimator depends on the ratio between the true reward and the probability, while that of the DR estimator depends on the ratio between the model error ($\re(x)-\model(x))$ and the sampling probability. Thus, the variance of the IPW estimator is uniquely zero if we sample proportionally to the true reward, while the variance of the DR estimator is zero if we sample according to the model residuals.

Observe that the variance of both $\htotipw$ and $\htotdr$ depend on the square of $1/\pi(x)$. It is this relationship which induces the trade off between reward and variance: As we place higher probability on those observations with high expected reward, the probability of other observations decreases proportionally and variance increases.

\section{Methods}
\subsection{Optimization Objective}
A natural approach to minimizing an estimator's variance while maximizing reward is to form a linear combination of the two objectives. For any set $X\subset\X$, consider the optimization problem
\begin{equation}
\label{eq:variance_objective}
    \sup_{\bpi\in\Pi_K(X)} \Phi_\beta(\bpi) = \E_\pi[\reward(\pi,\model)] - \beta \Var_\pi(\htot),
\end{equation}
which selects inclusion probabilities in order to maximize reward and minimize variance. The trade-off between the two objectives is controlled by a predetermined scalar $\beta\in\R_{\geq 0}$.   
Higher values of $\beta$ place more emphasis on variance minimization. 
The set of legal inclusion probabilities over which the optimization takes place is 
\begin{equation*}
    \Pi_K(X) = \bigg\{\pi\in (0,1]^X:  \sum_{x\in X} \pi(x) =K\bigg\}, 
\end{equation*}
which requires that probabilities be strictly greater than zero to ensure that $\htotipw$ and $\htotdr$ are well-defined. Here, 
\begin{equation*}
    \E_\pi[\reward(\pi,\model)] = \sum_x \pi(x) \model(x),
\end{equation*}
is the expected reward according to the model. We note that the supremum is required in \eqref{eq:variance_objective} since $\Pi_K(X)$ is not closed. 

Unfortunately, if $\htot$ is the IPW or the DR estimator, Equation \eqref{eq:variance_objective} leads to a rather intractable optimization problem. We provide a more thorough discussion of the difficulties in appendix~\ref{app:intractable}, but suffice it to say that the joint inclusion probabilities $\pi(x,z)$ in Equation~\eqref{eq:general_variance} do not readily lend themselves to optimization. Indeed, for most batched sampling strategies, they do not have closed form solutions. 

\subsection{Model-Proportional Sampling}

Despite being difficult to optimize directly, for the IPW estimator $\htotipw$, the optimization problem~\eqref{eq:variance_objective} has an interesting property. Notice that $A_T(\theta)=0$ in Eq.~\ref{eq:general_variance} if $\theta(x)/\pi(x)=C$ for some constant $C$ for all $x$. Thus, if using $\htotipw$, a natural strategy is to sample according to 
\begin{equation}
    \label{eq:prop_probs}
    \pi(x) = \frac{K\model(x)}{\sum_z\model(z)},
\end{equation}
if $K\model(x)/\sum_z\model(z)\leq 1$. We call this approach \modelprop (\mps).  
There are several drawbacks to this approach. The first is that it relies on the model $\model$. If the model error is large, then so too will be the variance. This method also provides no way to tradeoff between variance and reward; its only focus is on minimizing variance. Indeed, we'll see in the results section that while the variance is low (if model error is reasonable), reward is also much lower than other methods. The next section uses KL-divergence to generalize this approach, enabling us to smoothly transition between probabilities in \eqref{eq:prop_probs} and those which place more weight on expected reward.

\subsection{Entropy and KL Sampling}
\label{sec:entropy_and_kl}

Due to the difficulties imposed by the variance term in objective function~\eqref{eq:variance_objective}, we propose two new optimization problems which are analytically tractable while still enabling a trade-off between variance and expected reward. The first uses entropy~\citep{shannon1948mathematical}. Fix a timestep $t$, and consider the sequential version of the problem (i.e., $K=1$). The \emph{entropy} of the sample $S=S_t$ (the random variable describing which observations are sampled), is 
\begin{equation}
H(S) = -\sum_{x\in X}\pi(x)\log(\pi(x)).     
\end{equation}

$H(S)$ is, roughly, a measure of how spread out the distribution $\pi$ is. 
As $H(S)$ increases, $\pi$ resembles a uniform distribution over $X$; as $H(S)$ decreases $\pi$ is skewed towards some subset of $X$. The second objective we'll consider is the Kullback–Leibler (KL) divergence~\citep{kullback1951information}. The KL divergence between two discrete distributions $P$ and $Q$ defined on the sample space $\Omega$ is  
\begin{equation*}
    \kld{P}{Q} = \sum_{\omega\in\Omega} P(\omega) \log(\frac{P(\omega)}{Q(\omega)}).
\end{equation*}
While the KL divergence has many interpretations depending on the application at hand, for our purposes we can think of it as measuring the ``difference'' between $P$ and $Q$. Minimizing the divergence as a function of $P$ pushes $P$ towards $Q$. 

Set $q(x)=\model(x)/\sum_z\model(z)$, i.e., the MPS sampling solution. From here, define two optimization problems over $\Pi(X)$:
\begin{align}
    \sup_{\pi\in\Pi_K(X)} \Phi^\ent_\beta(\pi) &= \E[\reward(\pi,\model)] + \beta H(S), \label{eq:ent_objective}  \\
    \sup_{\pi\in\Pi_K(X)} \Phi^\kl_\beta(\pi) &= \E[\reward(\pi,\model)] -\beta \kld{\pi}{q}, \label{eq:kl_objective} 
\end{align}

Unlike the RL setting, we do not assume that the parameters of $\Phi$ are regularized, but rather the sampling distribution itself. Due to the structure of the functions, these optimization problems can be analytically solved to find the following optimal sampling policies.

\begin{proposition}
\label{prop:opt}
For a set of observations $X\subset \X$ and  $\Pi_K(X)$ as above, the solutions to optimization problems \eqref{eq:ent_objective}, \eqref{eq:kl_objective} with $K=1$ are $\{\pi^\ent(x)\}_{x\in X}$ and $\{\pi^\kl(x)\}_{x\in X}$ respectively, where
\begin{equation}
    \label{eq:ent_solution}
    \pi^\ent(x) = \frac{\exp(\model(x)/\beta)}{\sum_z\exp(\model(z)/\beta)} = \softmax(\model(x)/\beta),
\end{equation}
\begin{equation}
    \label{eq:kl_solution}
    \pi^\kl(x) = \frac{\model(x)\exp(\model(x)/\beta)}{\sum_z \model(z) \exp(\model(z)/\beta)}.
\end{equation}
\end{proposition}

\subsection{Approximating $\pi$ for $K>1$}
\label{sec:pareto}

Equations~\eqref{eq:ent_objective} and \eqref{eq:kl_objective} are for the sequential setting. Unfortunately, solving the naive extension to the batched setting is intractable -- both analytically and computationally -- because the entropy of $S$ involves an exponentially large sum over all subsets of size $K$, and the calculation of the higher level inclusion probabilities:
\[H(S) = -\sum_{A\subset X, |A|=K}\pi_{A}\log\pi_A,\]
where $\pi_A=\Pr(A=S)$. 
Therefore, in order to scale up Equations~\eqref{eq:ent_solution} and \eqref{eq:kl_solution} to the batched setting, we begin by multiplying each $\pi(x)$ by $K$, i.e., $\pi^*(x)=K\textnormal{softmax}(\model(x)/\beta)$. Depending on the distribution of $\model(x)$, however, this quantity might be greater than 1. In this case, we set $\pi^*(x)=1$ (i.e., it is sampled deterministically), and recalculate the probabilities on the subpopulation for which $K\text{softmax}(\model(x)/\beta)<1$. This is repeated until no inclusion probabilities exceed 1. This is reflected in Algorithm~\ref{alg:entropy_sampling}.

\begin{algorithm}[t]
\caption{Entropy-regularized Pareto Sampling}
\label{alg:entropy_sampling}
\begin{minipage}{0.5\textwidth}
\begin{algorithmic}
\State $Z\gets X$
\State $\hpi(x)\gets K\frac{e^{\model(x)/\beta}}{\sum_{z\in Z}e^{\model(x)/\beta}}, \forall x \in Z$ (Eq.~\ref{eq:kl_solution} if KL)
\State $F\gets \emptyset$
\While{$\exists x: \hpi(x)>1$}
\State $F\gets F\cup \{x:\hpi(x)> 1\}$
\State $Z \gets X\setminus F$
\State $\hpi(x) \gets \frac{Ke^{\model(x)/\beta}}{\sum_{z\in Z}e^{\model(x)/\beta}}$, $\forall x\in Z$ (Eq.~\ref{eq:kl_solution} if KL)
\EndWhile
\State Sample $U(x)\sim \text{unif}(0,1)$, $\forall x \in X$
\State $V(x) \gets U(x)(1-\hpi(x)) / ((1-U(x))\hpi(x))$
\State Relabel s.t. $V(x_1)\leq V(x_2)\leq \dots \leq V(x_N)$
\State \Return $x_1,\dots,x_K$
\end{algorithmic}
\end{minipage}
\end{algorithm}

\begin{figure}[t]
    \centering
    \includegraphics[scale=0.4]{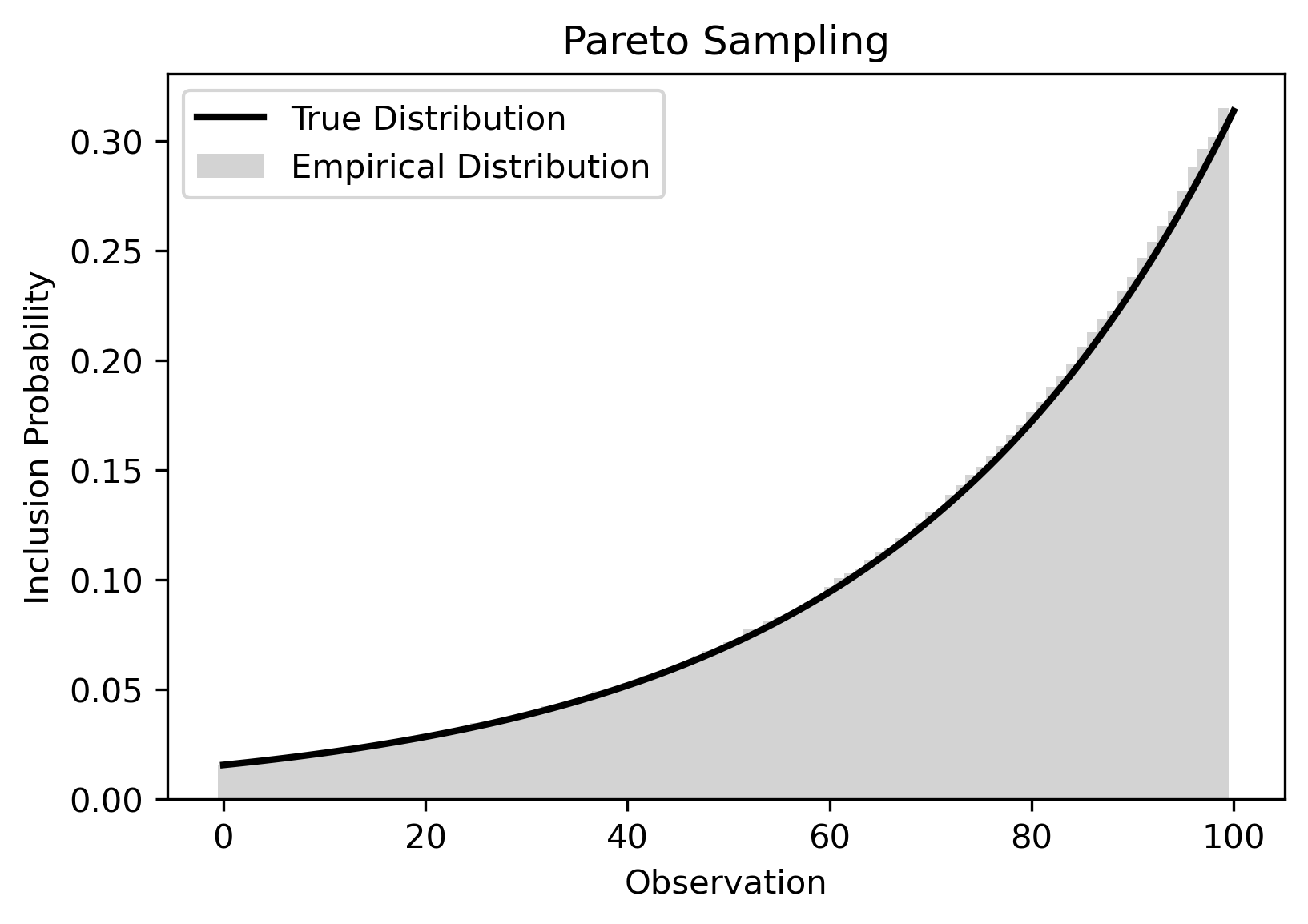}
    \caption{Illustration of Pareto sampling. We use $K=20$, and the true distribution was generated as $p(x) = 20\cdot\softmax(3x)$ for $x\in\{0, 0.01, 0.02, \dots, 1\}$. Pareto sampling was performed 10,000 times to obtain the empirical inclusion probability. }
    \label{fig:pareto}
\end{figure}

Once the inclusion probabilities are computed, the agent must actually sample from that distribution. This is trivial in the sequential setting, but more complicated in the batched setting. Designing a sampling scheme which respects first order inclusion probabilities precisely is difficult. Sampford sampling \citep{sampford1967sampling}, for instance, can guarantee prespecified first order inclusion probabilities, but is infeasible as sample sizes become large as it is a rejective procedure.  Instead, we employ Pareto Sampling~\citep{rosen1997sampling}. Here, given $N$ target inclusion probabilities $\hpi(x)$, we generate $N$ random values 
\[V(x) = \frac{U(x)(1-\hpi(x))}{(1-U(x))\hpi(x)},\quad U(x)\sim\text{unif}(0,1).\]
The $K$ samples with the smallest values are selected. This method is fast and always yields a sample of size $K$. The drawback is that the method is only approximate: The true inclusion probabilities $\pi$ are not precisely equal to $\hpi$. However, \citet{rosen2000inclusion} showed that the approximation error goes to zero as $K$ increases: 
\[\max_x |\pi(x)/\hpi(x)-1|=O(\log K /\sqrt{K}).\] 
Moreover, in practice the method works extremely well. See Figure~\ref{fig:pareto} for an illustration of its accuracy even at the relatively low budget of $K=20$. We note that, as \citet{rosen2000inclusion} discusses, in practice any bias introduced by this approximation is empirically low. And this can be further reduced by calculating exact inclusion probabilities through numerical means at the cost of time.

\section{Variance Bounds}
\label{sec:theory}

In this section we study the variance of $\htotipw$ and $\htotdr$ under Entropy and KL Sampling. Throughout, we fix a time $T$ and condition on the previous observations and model choices, in addition to the population draw from $\D$. Thus, the only randomness stems from the sampling itself. 
Moreover, we focus on the set of observations which will not be sampled with certainty. This is captured by the following assumption.

\begin{assumption}
\label{asump:Kpi}
For all $x\in\X$, $K\pi^\ent(x)\leq 1$ and $K\pi^\kl(x)\leq 1$ where $\pi^\ent$ and $\pi^\kl$ are as in Equations~\eqref{eq:ent_solution} and \eqref{eq:kl_solution}.  
\end{assumption}

We provide upper bounds on the variance which do not involve the joint inclusion probabilities. This is useful, because they can be calculated \emph{a priori} using only the model predictions. For a given period $T$ and set of observations $X_T$, let $\modelmin=\min_{x\in X_T}\model(x)$. 

\begin{proposition}[Variance Bound for Entropy Sampling]
\label{prop:var_bound_entropy}
Let $g_x=\exp(\model(x)/\beta - \modelmin/\beta)$ the $\beta$-weighted gap between the model prediction for $x$ and the minimum prediction in exponential space. Define 
\begin{equation*}
    C^\ent_1 = \frac{1}{K}\sum_{x,z}g_xg_z - \sum_x g_x^2.
\end{equation*}
If we sample precisely according $\pi^\ent$ (Equation~\eqref{eq:ent_solution}) (i.e., $\hpi(x)=\pi(x)$) then Entropy sampling obeys  $\Var(\htotipw(T)) \leq C^\ent_1$ and $\Var(\htotdr(T))\leq 2C^\ent_1$. 
\end{proposition}

\begin{figure}[t]
    \centering
    \begin{minipage}{0.38\textwidth}
    \includegraphics[scale=0.5]{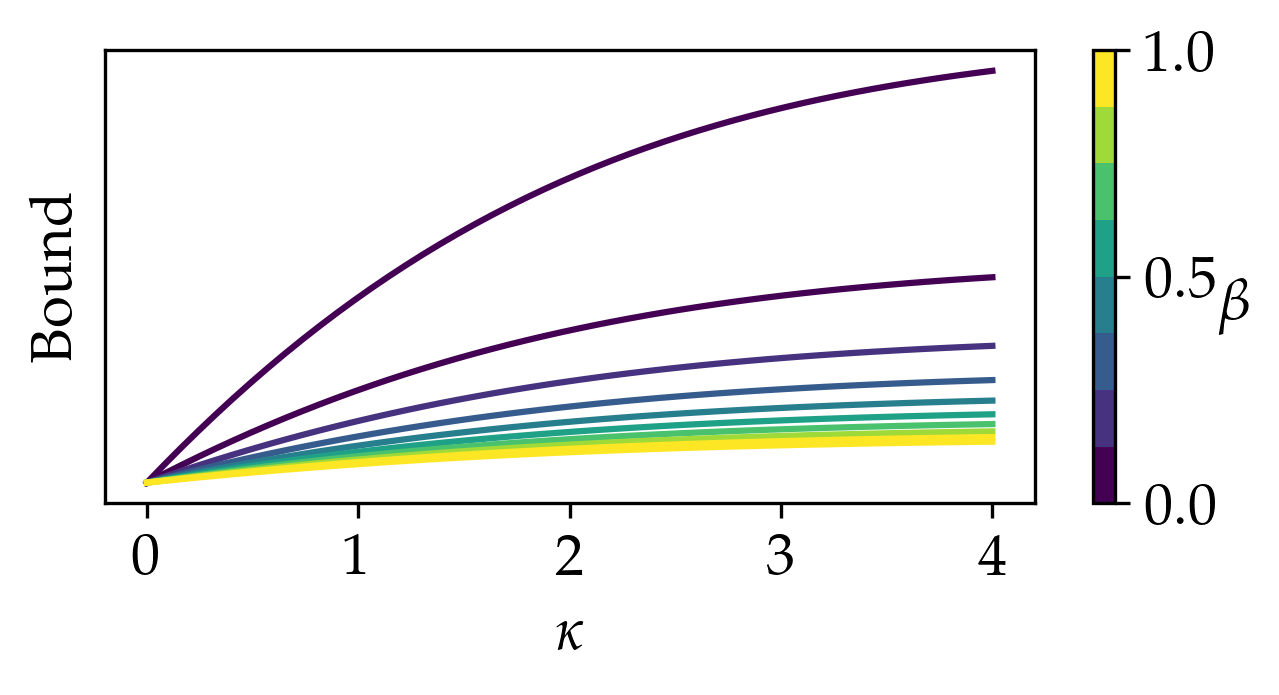}
    \end{minipage}
    \begin{minipage}{0.38\textwidth}
    \includegraphics[scale=0.5]{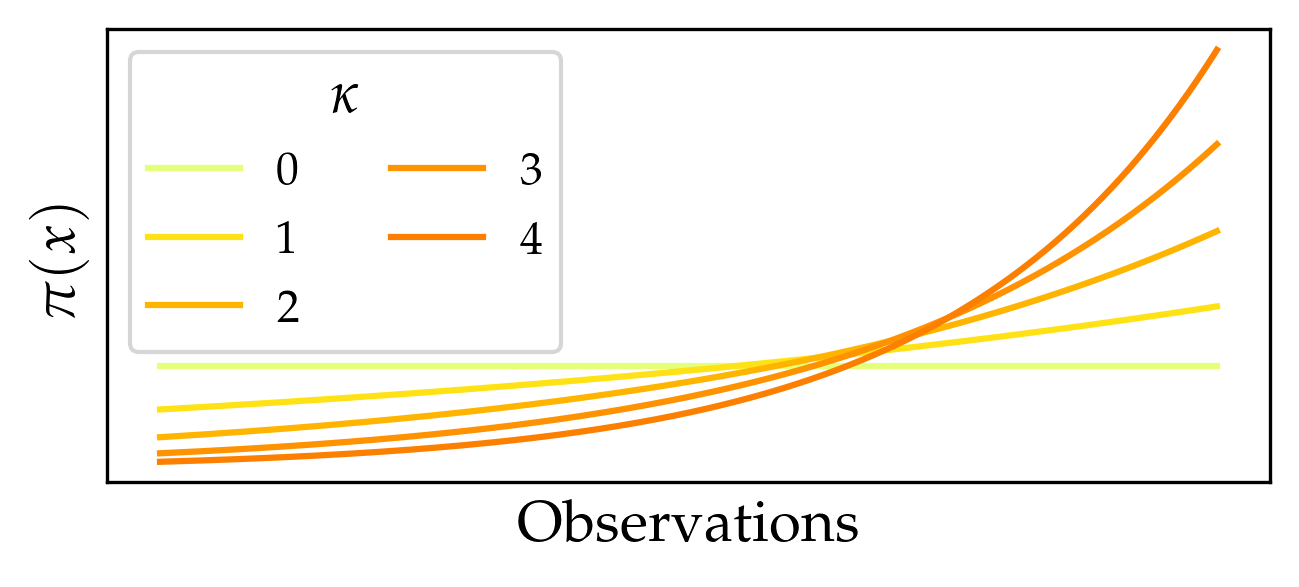}
    \vspace{0.3cm}
    \end{minipage}
    \caption{Illustration of Proposition~\ref{prop:var_bound_entropy}. Left: Change in variance bound as a function of $\beta$ and the shape of the inclusion probability distribution as characterized by $\kappa$. The effect of $\kappa$ on the inclusion probabilities is shown by the rightmost figure. }
    \label{fig:var_bound}
\end{figure}

Figure~\ref{fig:var_bound} demonstrates the bound for various values of $\beta$ and shapes of the inclusion probability distribution $\pi(x)$. We find that the value of $\beta$ has a much greater effect on the bound than the shape of the distribution. The following proposition gives the equivalent bound for KL sampling. 

\begin{proposition}[Variance Bound for KL Sampling]
\label{prop:var_bound_kl}
Let $g_x$ be as in Proposition~\ref{prop:var_bound_entropy}, and define 
\begin{equation*}
    C_1^\kl = \frac{1}{K}\sum_{x,z}\frac{\model_x\model_z}{\modelmin^2}g_xg_z - \sum_x \frac{\model^2_x}{\modelmin^2}g_x^2.
\end{equation*}
If we sample precisely according $\pi^\kl$ (Equation~\eqref{eq:kl_solution}) (i.e., $\hpi(x)=\pi(x)$) then KL sampling obeys 
$\Var(\htotipw(T))\leq C_1^\kl$ and $\Var(\htotdr(T))\leq 2C_1^\kl$. 
\end{proposition}

Beyond explicit calculations, an alternative way to grapple with the variance is to consider repeated trials of the sampling procedure. The following proposition thus calculates concentration bounds for the IPW estimator. While the result contains the typical quantities found in a Hoeffding-like bound, the more interesting term is
\begin{equation}
\label{eq:gamma}
  \Gamma^\A(\model) = \sum_{x\in X_T}D_x(\A) \exp(\frac{1}{\beta}(\model(x)-\modelmin)),
\end{equation} 
where $D_x(\A) = 1$ for $\A=$ Entropy Sampling and $D_x(\A) = \model(x)/\modelmin$ for $\A=$ KL Sampling. 
This may be interpreted as measure of the skew of the set of model predictions. Indeed, $\Gamma^\A(\model)$ is minimized when $\model(x)=\modelmin$ for all $x$.

\begin{proposition}(Hoeffding Bound for IPW sampling)
\label{prop:hoeffding_ipw}
At round $T$, suppose we repeat the sampling procedure $m$ times, yielding estimates $\htotipw^1,\dots,\htotipw^m$. If we sample according precisely according to the inclusion probabilities, then for any $\A\in\{\text{Entropy Sampling, KL Sampling}\}$, then for any $\delta>0$, with probability at least $1-\delta$, 
\begin{equation}
\label{eq:hoeff_precise}
    \bigg|\frac{1}{m}\sum_{i=1}^m \htotipw^i(T) - \tot\bigg| < \frac{\Gamma^\A(\model)}{\sqrt{2m}}\log^{1/2}\bigg(\frac{2}{\delta}\bigg),
\end{equation}
where $\Gamma^\A(\model)$ is as in Equation~\eqref{eq:gamma}. If we use Pareto sampling to approximate the inclusion probabilities, then with the same probability,
\begin{align}
    \bigg|\frac{1}{m}\sum_{i=1}^m \htotipw^i(T) - \tot\bigg| &< \frac{\Gamma^\A(\model)}{\sqrt{2m}}\log^{1/2}\bigg(\frac{2}{\delta}\bigg) + O(\log K /\sqrt{K})\tot \label{eq:hoeff_pareto} \\ 
    &\leq \frac{\Gamma^\A(\model)}{\sqrt{2m}}\log^{1/2}\bigg(\frac{2}{\delta}\bigg) + O(N \log K /\sqrt{K}). \label{eq:hoeff_pareto_N}
\end{align}
\end{proposition}

Proposition~\ref{prop:hoeffding_ipw} makes clear the dependence on $\beta$: As $\beta$ increases, $\Gamma(\model)$ decreases, and the bound grows tighter. 
We note that Equation~\eqref{eq:hoeff_pareto} is a form of relative error, since $\tot$ appears on the right hand side. This can be alleviated at the cost of a looser bound (Equation~\eqref{eq:hoeff_pareto_N}), by noting that $\tot\leq N$ for rewards bounded in $[0,1]$. 
Unfortunately, it is difficult to further tighten the gap when using Pareto Sampling, owing to the loose error rate of $O(\log K/\sqrt{K})$. As we saw in Section~\ref{sec:pareto}, the empirical error is much lower, but obtaining tighter theoretical bounds has proven difficult.

\section{Experiments}
\label{sec:experiments}

\subsection{Datasets}

We run experiments on four publicly available datasets: The Current Population Survey (CPS), the American Community Survey (ACS), a voter turnout dataset, and data on AllState severity claims. These four were chosen because they each correspond to a real-world optimize-and-estimate setting. For instance, the CPS dataset is closely related to the tax-gap estimation done by the IRS each year~\citep{henderson2022integrating}. The AllState population estimation task corresponds to an insurance company which must estimate the average cost of claims across the population. More detail on each dataset and further justification for their selection can be found in Appendix~\ref{app:datasets}.

\subsection{Baselines}

We compare entropy sampling to both \mps (described above), and also to simple random sampling (\srs). \srs calculates a population estimate based on the sampled rewards, i.e., it does not use a model. We also use Adaptive Bin Sampling (ABS), introduced by \citet{henderson2022integrating}. 
A full overview of ABS is given in Appendix~\ref{app:abs}, but we provide a brief treatment here. 
Each period, ABS parameterizes the model predictions with either a logistic function or an exponential function. This process is called ``smoothing'', and is controlled by a parameter $\alpha$ which dictates the shape of the paremarization. As such, it controls the reward-variance tradeoff for ABS: as $\alpha\to\infty$ more weight is placed on observations with higher predicted reward, and as $\alpha\to0$ we approach random sampling. The actual sampling mechanism is a two-stage strategy, which clusters predictions with similar (parameterized) predicted rewards, samples the clusters according to their average reward value, and then samples uniformly at random within clusters. This strategy enables ABS to calculate precise inclusion probabilities, thereby avoiding the need for pareto sampling. The drawback is the lack of observation-specific inclusion probabilities, as each observation in the same cluster shares the same probability. Moreover, clusters must have a minimum size of $K$, placing a limit on how fine-grained the inclusion probabilities can be.  While the authors of ABS use only the IPW estimator in their experiments, we test ABS with all three estimators: IPW, DR, and Model-based.

\subsection{Experimental Protocol}

For each dataset and method, observations for the first period are selected uniformly at random to provide a initial training set for the model. Because we're interested primarily in the performance of the sampling methods themselves, we hold the model constant across sampling algorithms and datasets. Due to the relatively small budget sizes and some evidence that tree-based methods outperform neural networks on tabular data~\citep{grinsztajn2022tree}, we use random forest regressors. 
We perform a randomized grid search on a small holdout set to determine a suitable set of hyperparameters for each dataset (see Appendix~\ref{app:tuning} for more details). 
Throughout our experiments, we keep the budget between approximately 1-10\% of the dataset size in each period, i.e., $K_t\in[0.01, 0.1]X_t$. We study the effect of budget size in more detail in Appendix~\ref{app:results}, but find that results remain consistent.

\subsection{Results}
\label{sec:results}

\begin{figure*}[ht!]
    \centering
    \includegraphics[scale=0.4]{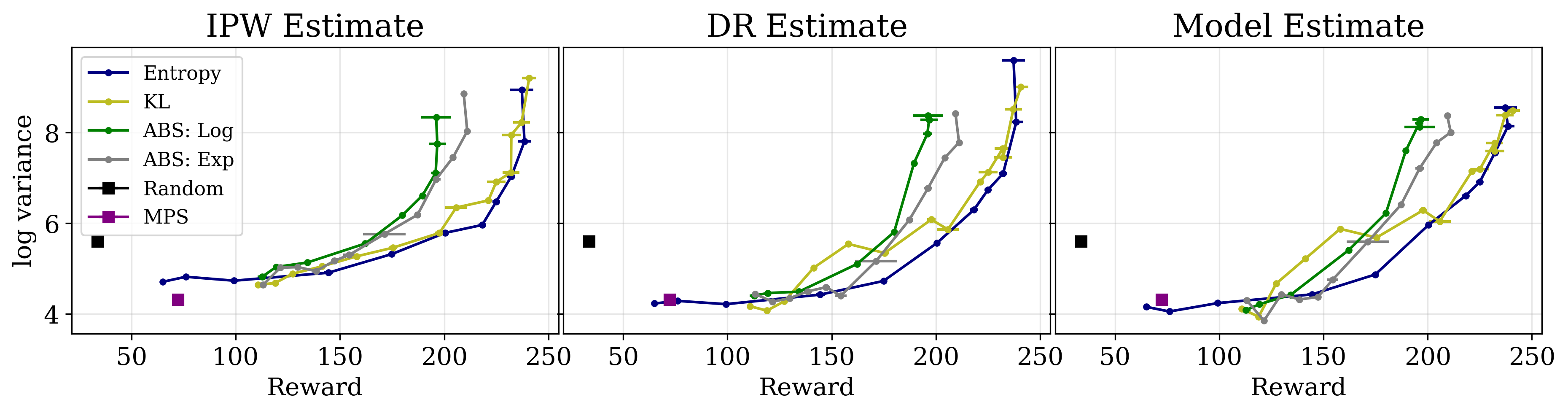}
    \includegraphics[scale=0.4]{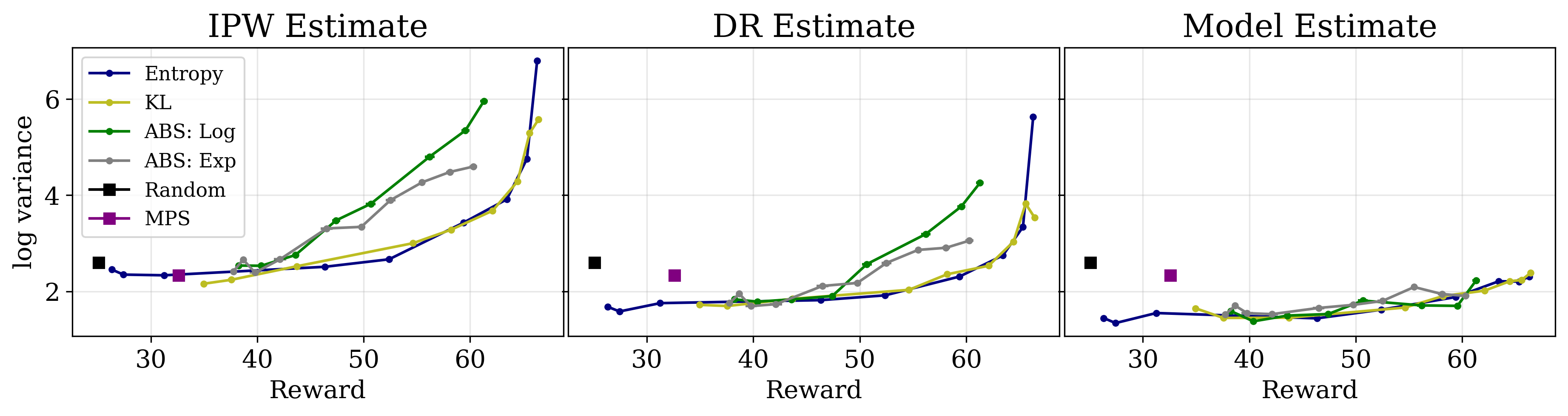}
    \caption{Reward-variance curves for all three estimators on the ACS dataset (top) and the AllState dataset (bottom). We used a budget of 1000 for both. Experiments were repeated at least 50 times to obtain errorbars (95\% CI) on reward. ABS: Log and ABS: Exp correspond to ABS with logistic and exponential smoothing, respectively. The scale of the $y$-axis is held constant for each dataset to keep variance in perspective across estimators.
    }
    \label{fig:reward_variance}
\end{figure*}

The full suite of experimental results as well as further discussion can be found in Appendix~\ref{app:results}. There, we demonstrate that results are consistent across datasets (besides those we highlight here), and across time steps. Here, we distill the results into four main takeaways. Due to the consistency across datasets, we we do not show figures for each one. To avoid selection bias, all figures in this section are plotted using the results from the final period.

\paragraph{KL and Entropy Samping improve the Reward-Variance Tradeoff.} As illustrated by Figure~\ref{fig:reward_variance}, both KL and Entropy sampling (mostly) improve over ABS in terms of the reward-variance trade off, especially for IPW and DR estimators. In fact, both are pareto improvements over ABS except occasionally in the low-reward region. 
For model estimation, KL and Entropy are never worse than ABS, and sometimes better.
These trends hold across all datasets. Interestingly, Entropy and KL Sampling perform comparably, except perhaps for the fact that Entropy Sampling tends to ride the variance-reward curve more smoothly. We conjecture that this is due to form of $\pi^\kl$ (Equation~\ref{eq:kl_objective}), which has added multiplicative terms dependent on the model outcomes as compared to the softmax. 

\begin{figure}[t]
    \centering
    \begin{minipage}{0.4\textwidth}
    \subcaptionbox{Bias of Entropy Sampling.\label{fig:error_bias_entropy}}{\includegraphics[scale=0.4,center]{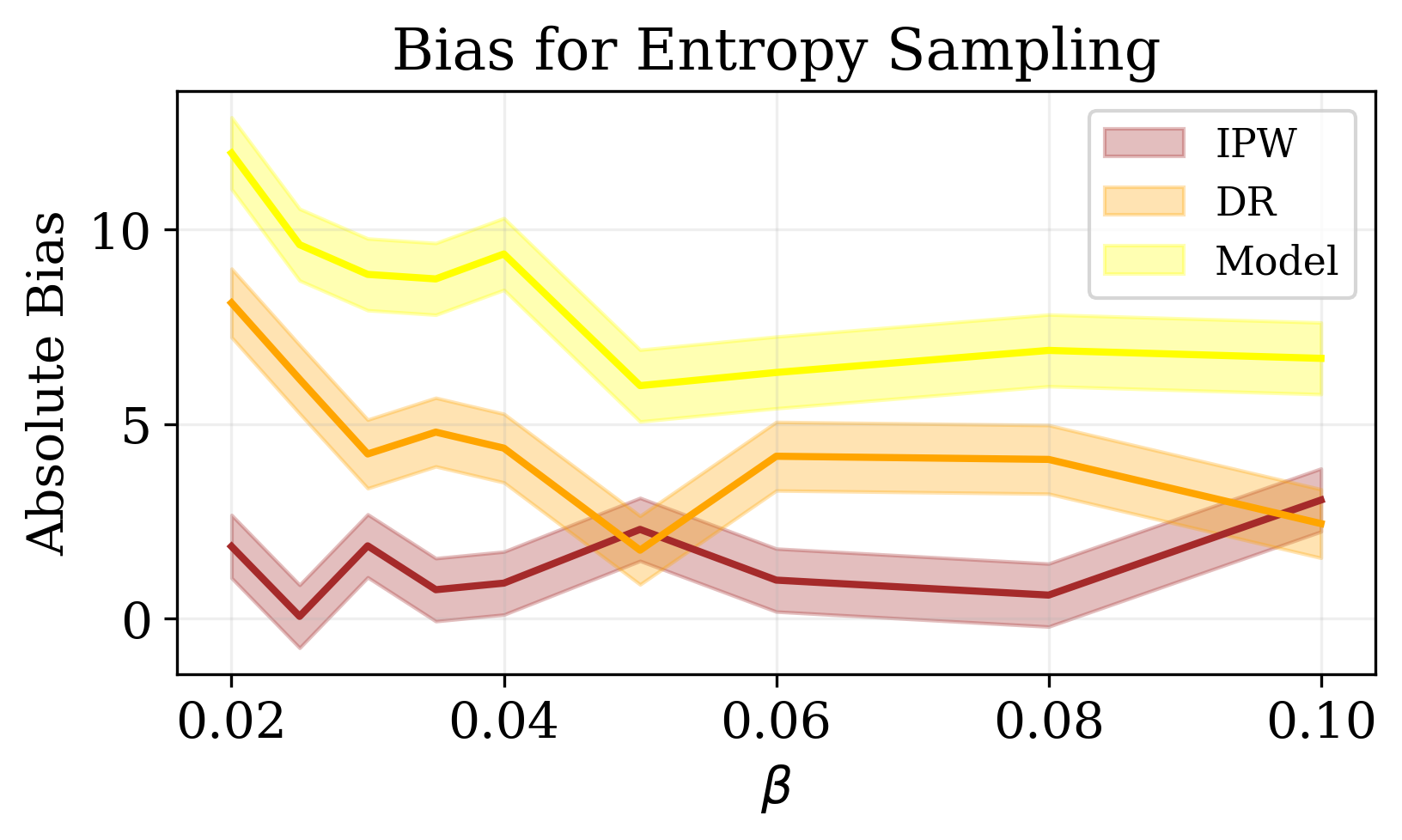}}
    \end{minipage}
    \begin{minipage}{0.4\textwidth}
    \subcaptionbox{Bias of IPW estimator. \label{fig:bias_ipw}}{\includegraphics[scale=0.4,center]{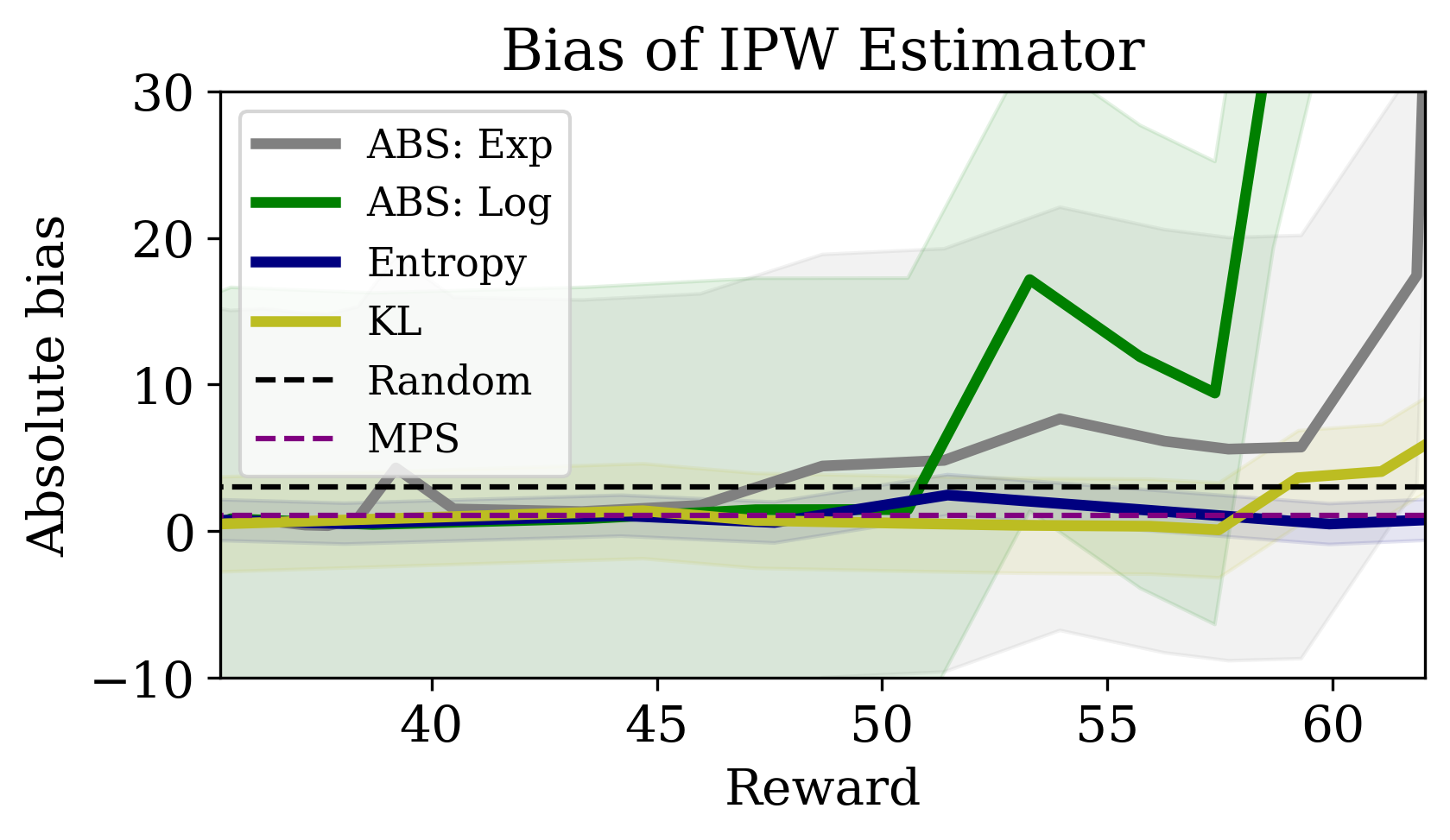}}
    \end{minipage}
    \caption{Left: Bias (with 95\% CI bands) plotted across values of $\beta$ for Entropy Sampling on the CPS dataset. Right: Absolute bias (with 95\% CI bands) of IPW estimator on CPS data as a function of reward. The x-axis is clipped to the largest reward range covered by all algorithms. }
\end{figure}

\paragraph{DR is more biased than IPW but less biased than model estimation.} As testified by Lemma~\ref{lem:bias}, all three estimators exhibit some form of bias if the model is mis-specified or the inclusion probabilities are only approximate. Figure~\ref{fig:error_bias_entropy} demonstrates the bias of entropy sampling on the ACS data, which is representative of the result across all datasets. Reliably, model estimation is the most biased, IPW the least, and DR between them. This is because DR is affected by model error (given inexact inclusion probabilities), while IPW is not. 

\paragraph{Model estimation has unreliable variance.} 
On the CPS, Voting, and AllState data, the variance of the model estimate hardly changes as a function of reward (Figure~\ref{fig:reward_variance} and Figure~\ref{fig:reward_variance_all_datasets} in Appendix~\ref{app:results}). For ACS data, on the other hand, the model estimates exhibit a similar reward-variance trade off to IPW and DR. This can be explained by model fit, which is significantly worse on ACS than on CPS. For instance, on CPS, the explained variance falls in the range [0.58, 0.64] across values of $\beta$ for Entropy Sampling. On ACS, it falls between [0.13, 0.7], a much wider range. Here, greater exploration of the model results in a much better fit, whereas on CPS, the difference is marginal.  Given a sufficiently well-fit model, the variance of the model estimate should be low, as the inclusion probabilities play no part in the estimator. The trade off, as seen in Figure~\ref{fig:error_bias_entropy}, is that model-estimates are more biased than either the IPW or DR estimator. In Appendix~\ref{app:results} we demonstrate that the variance of the model-estimate on ACS decreases as the budget increases.  

\paragraph{Entropy and KL Sampling maintain low bias with IPW.} 
Figure~\ref{fig:bias_ipw} illustrates the absolute bias of the IPW estimator across sampling strategies as a function of reward. Entropy and KL sampling both maintain low bias, even into the high-reward region where ABS begins to falter. This is a function of the reward-variance trade off of Figure~\ref{fig:reward_variance}: For the same reward, ABS yields higher variance, making bias more apparent unless averaged over a sufficient number of model runs. We note that the variance of ABS is much higher than Entropy and KL at the same reward level, consistent with Figure~\ref{fig:reward_variance}.  Figure~\ref{fig:bias_ipw} also testifies to the efficiency of the Pareto approximation. Indeed, the bias of entropy and KL rivals that of random and MPS.

\paragraph{MPS can be a Pareto improvement over Random Sampling.} MPS tends to have low variance but also low reward (though higher reward than random sampling, Figure~\ref{fig:reward_variance}). Surprisingly, it also has low bias while maintaining low variance, demonstrating the potential promise of model-based approaches in pure population estimation tasks.

\section{Conclusion}
We studied the optimize-and-estimate structured bandit setting, wherein the agent has the goal of not only maximizing their reward over time, but also of estimating the total reward over all arms in each period. 
We introduced two algorithms for this problem, both of which leverage policy regularization methods. The regularization terms (entropy and KL-divergence) allow for an analytical solution to the trade off between expected reward and estimator variance, thereby allowing practitioners to decide the relative importance they place on the two objectives. Experimentally, we find that the two algorithms improve upon the reward-variance trade off of current baselines, in addition to maintaining minimal bias of the population estimate. 

\subsection*{Ackowledgements}
We thank Dilip Arumugam, Brandon Anderson, Jia Wan for helpful feedback and discussions, and Vaden Masrani for shortening an otherwise overlong title. PH is supported by the Open Philanthropy AI Fellowship. This work was conducted while BC was at Stanford University.

\bibliographystyle{plainnat}
\bibliography{main.bib}

\begin{thebibliography}{44}
\providecommand{\natexlab}[1]{#1}
\providecommand{\url}[1]{\texttt{#1}}
\expandafter\ifx\csname urlstyle\endcsname\relax
  \providecommand{\doi}[1]{doi: #1}\else
  \providecommand{\doi}{doi: \begingroup \urlstyle{rm}\Url}\fi

\bibitem[Abbasi-Yadkori et~al.(2011)Abbasi-Yadkori, P{\'a}l, and
  Szepesv{\'a}ri]{abbasi2011improved}
Yasin Abbasi-Yadkori, D{\'a}vid P{\'a}l, and Csaba Szepesv{\'a}ri.
\newblock Improved algorithms for linear stochastic bandits.
\newblock \emph{Advances in neural information processing systems}, 24, 2011.

\bibitem[Ahmed et~al.(2019)Ahmed, Le~Roux, Norouzi, and
  Schuurmans]{ahmed2019understanding}
Zafarali Ahmed, Nicolas Le~Roux, Mohammad Norouzi, and Dale Schuurmans.
\newblock Understanding the impact of entropy on policy optimization.
\newblock In \emph{International conference on machine learning}, pages
  151--160. PMLR, 2019.

\bibitem[Aires(1999)]{aires1999algorithms}
Nibia Aires.
\newblock Algorithms to find exact inclusion probabilities for conditional
  poisson sampling and pareto $\pi$ps sampling designs.
\newblock \emph{Methodology and Computing in Applied Probability}, 1\penalty0
  (4):\penalty0 457--469, 1999.

\bibitem[Berger(1998)]{berger1998rate}
Yves~G Berger.
\newblock Rate of convergence for asymptotic variance of the horvitz--thompson
  estimator.
\newblock \emph{Journal of Statistical Planning and Inference}, 74\penalty0
  (1):\penalty0 149--168, 1998.

\bibitem[Bibaut et~al.(2021)Bibaut, Dimakopoulou, Kallus, Chambaz, and van~der
  Laan]{bibaut2021post}
Aur{\'e}lien Bibaut, Maria Dimakopoulou, Nathan Kallus, Antoine Chambaz, and
  Mark van~der Laan.
\newblock Post-contextual-bandit inference.
\newblock \emph{Advances in Neural Information Processing Systems},
  34:\penalty0 28548--28559, 2021.

\bibitem[Brekelmans et~al.(2022)Brekelmans, Genewein, Grau-Moya, Del{\'e}tang,
  Kunesch, Legg, and Ortega]{brekelmans2022your}
Rob Brekelmans, Tim Genewein, Jordi Grau-Moya, Gr{\'e}goire Del{\'e}tang,
  Markus Kunesch, Shane Legg, and Pedro Ortega.
\newblock Your policy regularizer is secretly an adversary.
\newblock \emph{arXiv preprint arXiv:2203.12592}, 2022.

\bibitem[Cassel et~al.(1976)Cassel, S{\"a}rndal, and Wretman]{cassel1976some}
Claes~M Cassel, Carl~E S{\"a}rndal, and Jan~H Wretman.
\newblock Some results on generalized difference estimation and generalized
  regression estimation for finite populations.
\newblock \emph{Biometrika}, 63\penalty0 (3):\penalty0 615--620, 1976.

\bibitem[Chugg and Ho(2021)]{chugg2021reconciling}
Ben Chugg and Daniel~E Ho.
\newblock Reconciling risk allocation and prevalence estimation in public
  health using batched bandits.
\newblock \emph{arXiv preprint arXiv:2110.13306}, 2021.

\bibitem[Deshpande et~al.(2018)Deshpande, Mackey, Syrgkanis, and
  Taddy]{deshpande2018accurate}
Yash Deshpande, Lester Mackey, Vasilis Syrgkanis, and Matt Taddy.
\newblock Accurate inference for adaptive linear models.
\newblock In \emph{International Conference on Machine Learning}, pages
  1194--1203. PMLR, 2018.

\bibitem[Deshpande et~al.(2019)Deshpande, Javanmard, and
  Mehrabi]{deshpande2019online}
Yash Deshpande, Adel Javanmard, and Mohammad Mehrabi.
\newblock Online debiasing for adaptively collected high-dimensional data.
\newblock \emph{arXiv preprint arXiv:1911.01040}, 2019.

\bibitem[Dimakopoulou et~al.(2019)Dimakopoulou, Zhou, Athey, and
  Imbens]{dimakopoulou2019balanced}
Maria Dimakopoulou, Zhengyuan Zhou, Susan Athey, and Guido Imbens.
\newblock Balanced linear contextual bandits.
\newblock In \emph{Proceedings of the AAAI Conference on Artificial
  Intelligence}, volume~33, pages 3445--3453, 2019.

\bibitem[Dimakopoulou et~al.(2021)Dimakopoulou, Ren, and
  Zhou]{dimakopoulou2021online}
Maria Dimakopoulou, Zhimei Ren, and Zhengyuan Zhou.
\newblock Online multi-armed bandits with adaptive inference.
\newblock \emph{Advances in Neural Information Processing Systems}, 34, 2021.

\bibitem[Drugan and Nowe(2013)]{drugan2013designing}
Madalina~M Drugan and Ann Nowe.
\newblock Designing multi-objective multi-armed bandits algorithms: A study.
\newblock In \emph{The 2013 International Joint Conference on Neural Networks
  (IJCNN)}, pages 1--8. IEEE, 2013.

\bibitem[Dud{\'\i}k et~al.(2011)Dud{\'\i}k, Langford, and Li]{dudik2011doubly}
Miroslav Dud{\'\i}k, John Langford, and Lihong Li.
\newblock Doubly robust policy evaluation and learning.
\newblock \emph{arXiv preprint arXiv:1103.4601}, 2011.

\bibitem[Dud{\'\i}k et~al.(2014)Dud{\'\i}k, Erhan, Langford, and
  Li]{dudik2014doubly}
Miroslav Dud{\'\i}k, Dumitru Erhan, John Langford, and Lihong Li.
\newblock Doubly robust policy evaluation and optimization.
\newblock \emph{Statistical Science}, 29\penalty0 (4):\penalty0 485--511, 2014.

\bibitem[Erraqabi et~al.(2017)Erraqabi, Lazaric, Valko, Brunskill, and
  Liu]{erraqabi2017trading}
Akram Erraqabi, Alessandro Lazaric, Michal Valko, Emma Brunskill, and Yun-En
  Liu.
\newblock Trading off rewards and errors in multi-armed bandits.
\newblock In \emph{Artificial Intelligence and Statistics}, pages 709--717.
  PMLR, 2017.

\bibitem[Fontaine et~al.(2019)Fontaine, Berthet, and
  Perchet]{fontaine2019regularized}
Xavier Fontaine, Quentin Berthet, and Vianney Perchet.
\newblock Regularized contextual bandits.
\newblock In \emph{The 22nd International Conference on Artificial Intelligence
  and Statistics}, pages 2144--2153. PMLR, 2019.

\bibitem[Gerber et~al.(2008)Gerber, Green, and Larimer]{gerber2008social}
Alan~S Gerber, Donald~P Green, and Christopher~W Larimer.
\newblock Social pressure and voter turnout: Evidence from a large-scale field
  experiment.
\newblock \emph{American political Science review}, 102\penalty0 (1):\penalty0
  33--48, 2008.

\bibitem[Grinsztajn et~al.(2022)Grinsztajn, Oyallon, and
  Varoquaux]{grinsztajn2022tree}
L{\'e}o Grinsztajn, Edouard Oyallon, and Ga{\"e}l Varoquaux.
\newblock Why do tree-based models still outperform deep learning on tabular
  data?
\newblock \emph{arXiv preprint arXiv:2207.08815}, 2022.

\bibitem[H{\'a}jek(1964)]{hajek1964asymptotic}
Jaroslav H{\'a}jek.
\newblock Asymptotic theory of rejective sampling with varying probabilities
  from a finite population.
\newblock \emph{The Annals of Mathematical Statistics}, 35\penalty0
  (4):\penalty0 1491--1523, 1964.

\bibitem[Henderson et~al.(2021)Henderson, Chugg, Anderson, and
  Ho]{henderson2021beyond}
Peter Henderson, Ben Chugg, Brandon Anderson, and Daniel~E Ho.
\newblock Beyond ads: Sequential decision-making algorithms in law and public
  policy.
\newblock \emph{arXiv preprint arXiv:2112.06833}, 2021.

\bibitem[Henderson et~al.(2022)Henderson, Chugg, Anderson, Altenburger, Turk,
  Guyton, Goldin, and Ho]{henderson2022integrating}
Peter Henderson, Ben Chugg, Brandon Anderson, Kristen Altenburger, Alex Turk,
  John Guyton, Jacob Goldin, and Daniel~E Ho.
\newblock Integrating reward maximization and population estimation: Sequential
  decision-making for internal revenue service audit selection.
\newblock \emph{arXiv preprint arXiv:2204.11910}, 2022.

\bibitem[Horvitz and Thompson(1952)]{horvitz1952generalization}
Daniel~G Horvitz and Donovan~J Thompson.
\newblock A generalization of sampling without replacement from a finite
  universe.
\newblock \emph{Journal of the American statistical Association}, 47\penalty0
  (260):\penalty0 663--685, 1952.

\bibitem[Jiang and Li(2016)]{jiang2016doubly}
Nan Jiang and Lihong Li.
\newblock Doubly robust off-policy value evaluation for reinforcement learning.
\newblock In \emph{International Conference on Machine Learning}, pages
  652--661. PMLR, 2016.

\bibitem[Joseph et~al.(2016)Joseph, Kearns, Morgenstern, Neel, and
  Roth]{joseph2016fair}
Matthew Joseph, Michael Kearns, Jamie Morgenstern, Seth Neel, and Aaron Roth.
\newblock Fair algorithms for infinite and contextual bandits.
\newblock \emph{arXiv preprint arXiv:1610.09559}, 2016.

\bibitem[Kullback and Leibler(1951)]{kullback1951information}
Solomon Kullback and Richard~A Leibler.
\newblock On information and sufficiency.
\newblock \emph{The annals of mathematical statistics}, 22\penalty0
  (1):\penalty0 79--86, 1951.

\bibitem[Narain(1951)]{narain1951sampling}
RD~Narain.
\newblock On sampling without replacement with varying probabilities.
\newblock \emph{Journal of the Indian Society of Agricultural Statistics},
  3\penalty0 (2):\penalty0 169--175, 1951.

\bibitem[Neel and Roth(2018)]{neel2018mitigating}
Seth Neel and Aaron Roth.
\newblock Mitigating bias in adaptive data gathering via differential privacy.
\newblock In \emph{International Conference on Machine Learning}, pages
  3720--3729. PMLR, 2018.

\bibitem[Nie et~al.(2018)Nie, Tian, Taylor, and Zou]{nie2018adaptively}
Xinkun Nie, Xiaoying Tian, Jonathan Taylor, and James Zou.
\newblock Why adaptively collected data have negative bias and how to correct
  for it.
\newblock In \emph{International Conference on Artificial Intelligence and
  Statistics}, pages 1261--1269. PMLR, 2018.

\bibitem[Ros{\'e}n(1997)]{rosen1997sampling}
Bengt Ros{\'e}n.
\newblock On sampling with probability proportional to size.
\newblock \emph{Journal of statistical planning and inference}, 62\penalty0
  (2):\penalty0 159--191, 1997.

\bibitem[Ros{\'e}n(2000)]{rosen2000inclusion}
Bengt Ros{\'e}n.
\newblock On inclusion probabilities for order $\pi$ps sampling.
\newblock \emph{Journal of statistical planning and inference}, 90\penalty0
  (1):\penalty0 117--143, 2000.

\bibitem[Russo and Zou(2016)]{russo2016controlling}
Daniel Russo and James Zou.
\newblock Controlling bias in adaptive data analysis using information theory.
\newblock In \emph{Artificial Intelligence and Statistics}, pages 1232--1240.
  PMLR, 2016.

\bibitem[Sampford(1967)]{sampford1967sampling}
Michael~R Sampford.
\newblock On sampling without replacement with unequal probabilities of
  selection.
\newblock \emph{Biometrika}, 54\penalty0 (3-4):\penalty0 499--513, 1967.

\bibitem[Shannon(1948)]{shannon1948mathematical}
Claude~Elwood Shannon.
\newblock A mathematical theory of communication.
\newblock \emph{The Bell system technical journal}, 27\penalty0 (3):\penalty0
  379--423, 1948.

\bibitem[Shin et~al.(2019)Shin, Ramdas, and Rinaldo]{shin2019sample}
Jaehyeok Shin, Aaditya Ramdas, and Alessandro Rinaldo.
\newblock Are sample means in multi-armed bandits positively or negatively
  biased?
\newblock \emph{Advances in Neural Information Processing Systems}, 32, 2019.

\bibitem[Shin et~al.(2020)Shin, Ramdas, and Rinaldo]{shin2020conditional}
Jaehyeok Shin, Aaditya Ramdas, and Alessandro Rinaldo.
\newblock On conditional versus marginal bias in multi-armed bandits.
\newblock In \emph{International Conference on Machine Learning}, pages
  8852--8861. PMLR, 2020.

\bibitem[Shin et~al.(2021)Shin, Ramdas, and Rinaldo]{shin2021bias}
Jaehyeok Shin, Aaditya Ramdas, and Alessandro Rinaldo.
\newblock On the bias, risk, and consistency of sample means in multi-armed
  bandits.
\newblock \emph{SIAM Journal on Mathematics of Data Science}, 3\penalty0
  (4):\penalty0 1278--1300, 2021.

\bibitem[Till{\'e}(2006)]{tille2006sampling}
Yves Till{\'e}.
\newblock \emph{Sampling algorithms}.
\newblock Springer, 2006.

\bibitem[Van~Moffaert et~al.(2014)Van~Moffaert, Van~Vaerenbergh, Vrancx, and
  Now{\'e}]{van2014multi}
Kristof Van~Moffaert, Kevin Van~Vaerenbergh, Peter Vrancx, and Ann Now{\'e}.
\newblock Multi-objective $\chi$-armed bandits.
\newblock In \emph{2014 International Joint Conference on Neural Networks
  (IJCNN)}, pages 2331--2338. IEEE, 2014.

\bibitem[Williams and Peng(1991)]{williams1991function}
Ronald~J Williams and Jing Peng.
\newblock Function optimization using connectionist reinforcement learning
  algorithms.
\newblock \emph{Connection Science}, 3\penalty0 (3):\penalty0 241--268, 1991.

\bibitem[Xiao et~al.(2019)Xiao, Huang, Mei, Schuurmans, and
  M{\"u}ller]{xiao2019maximum}
Chenjun Xiao, Ruitong Huang, Jincheng Mei, Dale Schuurmans, and Martin
  M{\"u}ller.
\newblock Maximum entropy monte-carlo planning.
\newblock \emph{Advances in Neural Information Processing Systems}, 32, 2019.

\bibitem[Yao et~al.(2021)Yao, Brunskill, Pan, Murphy, and
  Doshi-Velez]{yao2021power}
Jiayu Yao, Emma Brunskill, Weiwei Pan, Susan Murphy, and Finale Doshi-Velez.
\newblock Power constrained bandits.
\newblock In \emph{Machine Learning for Healthcare Conference}, pages 209--259.
  PMLR, 2021.

\bibitem[Zhang et~al.(2020)Zhang, Janson, and Murphy]{zhang2020inference}
Kelly Zhang, Lucas Janson, and Susan Murphy.
\newblock Inference for batched bandits.
\newblock \emph{Advances in neural information processing systems},
  33:\penalty0 9818--9829, 2020.

\bibitem[Zimmert and Seldin(2021)]{zimmert2021tsallis}
Julian Zimmert and Yevgeny Seldin.
\newblock Tsallis-inf: An optimal algorithm for stochastic and adversarial
  bandits.
\newblock \emph{J. Mach. Learn. Res.}, 22\penalty0 (28):\penalty0 1--49, 2021.

\end{thebibliography}

\appendix 
\onecolumn

\section{Extended Related Work}
\label{app:related_work}

\paragraph{Multi-objective optimization and multi-objective and bandits.}
Aside from the optimize-and-estimate setting specifically, the dominant paradigm for multi-objective in online learning involves considering rewards to be vectors instead of scalars, and searching for solutions on the pareto front~\citep{drugan2013designing,van2014multi}. 
Typically, an ``aggregation'' or ``scalarizing'' function is used to collapse the vector into a scalar in order to perform traditional optimization. 
With respect to bandits in particular, \citet{erraqabi2017trading} study how to trade off reward and model accuracy in multi-armed bandits. Like us, they formulate the problem as a linear combination of the two objectives. Unlike us, however, they do not work in the contextual setting, and the set of arms are not changing over time. \citet{yao2021power} study the joint problem of reward maximization and hypothesis testing under semi-parametric bandits. They trade-off between reward and sampling with sufficient statistical power.

\paragraph{Reliable ex-post inference.}
\citet{deshpande2019online,deshpande2018accurate} propose a method to perform reliable ex post inference with adaptively collected data assuming a linear model. \citet{dimakopoulou2019balanced} introduced balanced contextual bandits, which use inverse propensity weighting to correct model estimates when rewards are linear.  \citet{bibaut2021post} introduce an asymptotically normal estimator in contextual bandit settings. 

\paragraph{Adaptive Bin Sampling.}
As mentioned previously, Henderson et al. introduce \emph{Adaptive Bin Sampling} (ABS), which provides an unbiased (in expectation) population estimate. Roughly speaking, ABS assigns a probability of selection to groups of observations based on a smoothed distribution of the model predictions and applies inverse propensity weighting (IPW)~\cite{horvitz1952generalization,narain1951sampling} to obtain its estimate. We give more detail on ABS in Appendix~\ref{app:abs}. As it is the only method to date explicitly designed for the optimize-and-estimate setting, it's worth discussing some of the differences between Entropy and KL Sampling and ABS.   

First, while both approaches serve to assign each observation a probability of inclusion in the sample, our approach does this by solving a convex optimization problem which enables us to explicitly write down the solution as a function of the model predictions. This lends itself much more readily to theoretical analysis than ABS, which relies on several steps such as K-Means clustering, making an explicit calculation of the eventual inclusion probabilities impossible. Second, ABS clusters observations such that all points in the same cluster have the same inclusion probability. There are pros and cons to this approach. Clustering enables a two-stage sampling strategy -- sample clusters with replacement, and within clusters without. This allows the agent to sample precisely according to given inclusion probabilities. Our approach, meanwhile, enables observation-level inclusion probabilities, but must employ approximate sampling methods (Pareto Sampling, Section~\ref{sec:pareto}).  
Third ABS requires more hyperparameters such as the number of clusters and the number of greedy selections to make. Our approach, meanwhile, enables observation level probabilities and requires fewer parameters. Fourth, while both approaches use a parameter to trade off between expected reward and variance -- $\alpha$ for ABS and $\beta$ for us -- this parameter can be chosen beforehand in our method, whereas $\alpha$ is difficult to calibrate prior to experiments due to the nature of the algorithm.  Finally, unlike the authors of ABS, we investigate the doubly-robust estimator~\citet{dudik2014doubly} in addition to the IPW estimator.

\paragraph{Off Policy Evaluation and Optimal Exploration.}
Entropy regularization is a popular tool in reinforcement learning, where it serves to encourage exploration in policy optimization~\citep{williams1991function, zimmert2021tsallis,ahmed2019understanding}.  \citet{fontaine2019regularized} employ entropy and KL-divergence in multi-armed bandits as regularization tools. While the problem settings are different, we will employ some of their tools in, e.g., KL-Sampling, in which we try to keep our selection policy close to MPS. Similarly to our work, they impose no assumptions to the reward function of each arm beyond basic regularity conditions. Of even closer relation to our setting is recent work by \citet{xiao2019maximum} in maximum entropy Monte-Carlo planning. Like us, they explicitly consider an objective function which is a linear combination of expected reward and entropy. However, they do not consider the KL variant, and do not consider population estimation and are not in a structured setting.

\section{Datasets}
\label{app:datasets}

\paragraph{Current Population Survey (CPS), 2018.} The ASEC data is a supplement to the Current Population Survey (CPS), and provides the official household income statistics for the US government. We use the 2019 ASEC data, which is comprised of 94,633 observations. We use the total positive income of the respondent as the reward (coded as `HTOTVAL'). We remove five features which have high correlation (above 0.4) with the reward. This leaves us with 122 features. We  split the dataset into 8 periods by grouping observations at random. This gives approximately 12K observations per period. Here, the population estimate refers to the estimating the average income of survey respondents. 

\paragraph{American Community Survey (ACS), 5 Year -- 2014-2018.} These data are a subset of the responses to the American Community Survey (ACS) from 2014-2018. We use each year as a distinct period. We use `PINCP' as the reward, which is the individual's total income. This is not inflation adjusted. We remove 11 features which had correlation of above 0.4 with the target, as well as all those features which are constant. This leaves us with 214 features, excluding the year and the reward. We replace all missing values with zeros. As above, the population estimate for the ACS data corresponds to the average total income for each individual. 

\paragraph{Voter Turnout.} The third dataset is from a field experiment on social pressure and voter turnout by \citet{gerber2008social}. There were four treatments designed to increase voter turnout, each of which was administered to 20,000 voters. We let the first period be a set of 20,000 voters in the  control group, and the final period be the remainder of the control group. This gives us six periods. There were 180,003 observations in total. Unlike the other datasets, these labels are binary. Here, the population estimate is an estimate of the fraction of voters who voted when administered the given treatment. Note that this is not quite the average treatment effect, since it does not contain the estimate of the control group. 

\paragraph{AllState Severity Claims.} This is an anonymized dataset which includes information on insurance claims. The goal is to predict their cost. There are 188,813 observations and 130 features, of which 116 are categorical. We first trim down the dataset and keep only the first 30 categorical variables. We apply one-hot-encoding to these variables for a final feature set size of 75.
We also randomly discard some observations, keeping a total of 113,210. Observations are then randomly partitioned into 6 periods. 
Here, the population estimate corresponds to estimating the average claim severity.

We chose these four datasets because each represents a feasible setting for  optimize-and-estimate bandits. CPS and ACS, for instance, represent the dual tasks of maximizing reward (income), while estimating the average income of the population. Population estimation in the voting dataset task corresponds to estimating the average voter turnout under the given treatment effect. We note that this is not equivalent to the average treatment effect since there's no control group, but this term could be added. The AllState dataset represents the task, common in insurance companies, of estimating the average claim.   

Together, the datasets provide several different perspectives on the sequential arrival of groups of observations. These perspectives are {\bf (a) random groupings} (CPS, AllState) --- data are drawn iid and from the same distribution at each period; {\bf (b) temporal groupings} (ACS) --- data in distinct periods were gathered at different times; and 
    {\bf (c) intervention groupings} (Voting) --- data in distinct periods were subject to distinct interventions (e.g., for an RCT).

\section{Details on Inclusion Probabilities}
\label{app:incl_probs}

There term ``probability'' is slightly misleading when applied to inclusion probabilities, because the set of numbers $\{\pi(x)\}_{x\in X}$ does not behave as a typical discrete probability distribution. Indeed, $\pi(x)$ describes not a single event, but a process of selecting $K$ elements. The underlying measure is one over the space $\S$ over all possible sets $S$ of $K$ elements of $X$. Let $\Pr_{\S}$ be such a measure. Inclusion probabilities are then the values \[\pi(x)=\Pr_\S(x\in S)=\sum_{S\subset X:|S|=K}\Pr(S)\delta_S(x),\] i.e., the probability that $x$ is including in the random sample $S$ which is drawn according to $\Pr_{\S}$. The joint inclusion probabilities are $\pi(x,z)=\Pr_\S(x,z\in S)$, and so on. 

The first intriguing feature of inclusion probabilities which betray the fact that they're not true probabilities is that they sum to $K$, not 1. Indeed, this should be obvious if one considers picking $K$ elements deterministically, in which case all of their inclusion probabilities are 1, and the rest are zero, giving a total sum of $K$. In general, we can write
\begin{equation}
    \sum_x \pi(x) = \sum_x \sum_S \Pr(S)\delta_S(x) = \sum_S \Pr(S)\sum_x \delta_S(x) = \sum_S \Pr(S) K = K,  
\end{equation}
since each valid sample $S$ has precisely $K$ elements (thus, note that this arithmetic does not work for sampling with replacement). Another feature worth noticing is the following. For fixed $x$, 
\begin{align*}
\sum_{z\neq x}\pi({x,z})&=\sum_{z\neq x}\sum_S \Pr(S)\delta_S(x)\delta_S(z)\\ 
&=\sum_{S:x\in S}\Pr(S)\sum_{z\neq x}\delta_S(z) \\ 
&= (K-1)\sum_{S:x\in S} \Pr(S) = (K-1)\pi(x),
\end{align*}
which enables us to prove a property useful in many of the variance calculations: 
\begin{align*}
  \sum_{z\neq x}(\pi(x)\pi(z) - \pi(x,z)) &= \pi(x)\sum_{z\neq x}\pi(z)  - (K-1)\pi(x)    \\
  &= \pi(x)\bigg(\sum_{z}\pi(z) - \pi(x)\bigg) - (K-1)\pi(x) \\
  &= \pi(x)(K-\pi(x)) - (K-1)\pi(x) 
  = \pi(x)(1-\pi(x)).
\end{align*}

\section{Omitted Proofs}
\subsection{Proof of Lemma~\ref{lem:bias}}
Write the model based estimator as 
\begin{equation*}
    \htotmod(T) = \sum_{x\in X_T}\bigg(r(x)\ind_{x\in S_T} + \model(x)(1-\ind_{x\in S_T})\bigg). 
\end{equation*}
Linearity of expectations gives 
\begin{align*}
    \E_\D\E_S[\htotmod(T)] &= \sum_{x\in X_T} \bigg(\E_\D[r(x)]\pi(x) + \model(x)(1-\pi(x)\bigg) \\
    &= \sum_{x\in X_T}\bigg((\E_D[r(x)] - \model(x))\pi(x) + \E_\D[r(x)] - \Delta(x)\bigg) \\
    &= \sum_{x\in X_T}\Delta(x)(\pi(x)-1) + \tot(T),
\end{align*}
from which the result immediately follows. For IPW we have 
\begin{equation*}
    \E_{\D,S}[\htotipw(T)] = \sum_{x\in X_T}\frac{\E_\D[r(x)]}{\hpi(x)}\E_S[\ind_{x\in S}] = \sum_{x\in X_T}\frac{\E_\D[r(x)]}{\hpi(x)}\pi(x) = \sum_{x\in X_T}\E_\D[r(x)]\lambda(x),
\end{equation*}
so, 
\begin{equation*}
    \E_{\D,S}[\htotipw(T)] - \tot =  \sum_{x\in X_T}\E_\D[r(x)]\lambda(x) - \sum_x \E[r(x)] = \sum_{x\in X_T}\E[r(x)](\lambda(x)-1).
\end{equation*}
For DR,  
\begin{align*}
    \E_{\D,S}[\htotdr(T)] &= \sum_{x\in X_T}\bigg(\E[\model(x)] + \frac{\E_\D[r(x)] - \model(x)}{\hpi(x)}\pi(x)\bigg) \\ 
    &= \sum_{x\in X_T} (\E_\D[r(x)] - \Delta(x) + \Delta(x)\lambda(x)) \\ 
    &= \tot(T) + \sum_{x\in X_T}\Delta(x)(\lambda(x)-1).
\end{align*}

\subsection{Proof of Proposition~\ref{prop:opt}}
\label{app:opt_soln} 

First consider the entropy objective. We can consider the closed set 
\[\bar{\Pi}(X) = \bigg\{\pi\in[N]^X: 0\leq \pi(x)\leq 1,\; \sum_{x\in X}\pi(x)=1\bigg\},\]
since, as we'll see, the solution does not fall on the boundary. The optimization problem can then be written as 
\[\max_{\pi} \Phi^\ent_\beta(\pi) = \E[\reward(\pi,\model)] + \beta H(S), \quad \text{s.t.} \quad 0\leq \pi(x)\leq 1, \;\sum_x\pi(x)=1.\]
This is a straightforward exercise in Lagrange multipliers. Consider, for the moment, the Lagrangian for the problem without the constraint $0\leq \pi\leq 1$ (but keeping the equality constraint): 
\[L(\pi,\lambda) = -\beta\sum_{x\in X}\pi(x)\log\pi(x) + \sum_x \pi(x)\model(x) - \lambda\bigg(\sum_x \pi(x) - 1\bigg).\]
Some arithmetic shows that $\partial L/\partial \pi(x)=0$ iff 
\begin{equation}
\label{eq:pi}
    \pi(x) = \exp\bigg(\frac{\model(x)-\lambda}{\beta}-1\bigg).
\end{equation}
The constraint $\sum_x\pi(x)=1$ implies that 
\[\sum_x \exp\bigg(\frac{\model(x)-\lambda}{\beta}-1\bigg) = \exp(-\lambda/\beta)\sum_x\exp(\frac{\model(x)}{\beta}-1) = 1,\]
i.e., 
\[\lambda = \beta\log\sum_x \exp(\frac{\model(x)}{\beta}-1).\]
Plugging $\lambda$ back into Equation~\eqref{eq:pi} and simplifying gives the result. A quick calculation shows that concavity is satisfied. Finally, notice that the solution trivially satisfies $0\leq \pi\leq1$. The other problem can be solved similarly. 

\subsection{Proof of Proposition~\ref{prop:var_bound_entropy}}

Recall that, 
\begin{equation*}
    \Var_\pi(\htot(T)) = \frac{1}{2}\sum_{x,z}\bigg(\frac{\theta(x)}{\pi(x)} - \frac{\theta(z)}{\pi(z)}\bigg)^2(\pi(x)\pi(z) - \pi(x,z)),
\end{equation*}
where $\theta$ takes a specific form for the IPW and DR estimators. Put 
\[A = \frac{1}{K}\sum_z\exp(\frac{\model(z)}{\beta}),\]
and note that for Entropy Sampling (and under Assumption 1),
\[\pi(x) = K\frac{e^{\model(x)/\beta}}{\sum_z e^{\model(z)/\beta}}.\]
We split the proof into two parts; one for the IPW estimator and one for the DR estimator. 

\paragraph{IPW Estimation.}
For IPW, $\theta(x)=r(x)$ which, by assumption, is bounded between 0 and 1. Therefore, 
\begin{align*}
    \bigg(\frac{r(x)}{\pi(x)} - \frac{r(z)}{\pi(z)}\bigg)^2 &= A^2 \bigg(\frac{r(x)}{\exp(\model(x)/\beta)} - \frac{r(z)}{\exp(\model(z)/\beta)}\bigg)^2 \\ 
    &\leq A^2\bigg(r(x)^2\exp(-2\model(x)/\beta) + r(z)^2\exp(-2\model(z)/\beta)\bigg) \\ 
    &\leq 2A^2\max_x \bigg\{r(x)^2 \exp(-\frac{2\model(x)}{\beta})\bigg\} \\ 
    &\leq 2A^2\max_x \bigg\{ \exp(-\frac{2\model(x)}{\beta})\bigg\} \\ 
    &= 2A^2 \exp(-\frac{2\modelmin}{\beta}),
\end{align*} 
for $\modelmin=\min_x\model(x)$. Then, 
\begin{align*}
\Var(\htotipw) & = \frac{1}{2}\sum_{x,z}\bigg(\frac{r(x)}{\pi(x)} - \frac{r(z)}{\pi(z)}\bigg)^2(\pi(x)\pi(z)-\pi(x,z)) \\ 
&\leq A^2\exp(-2\modelmin/\beta) \sum_{x,z\neq x}(\pi(x)\pi(z) - \pi(x,z)).
\end{align*}
Employing the properties of inclusion probabilities in Appendix~\ref{app:incl_probs}, write
\begin{align*}
  \sum_{z\neq x}(\pi(x)\pi(z) - \pi(x,z)) &= \pi(x)\sum_{z\neq x}\pi(z)  - (K-1)\pi(x)   \\ 
  &= \pi(x)(K-\pi(x)) - (K-1)\pi(x) \\
  &= \pi(x)(1-\pi(x)),
\end{align*}
hence, 
\begin{align*}
    \sum_{x,z\neq x}(\pi(x)\pi(z)-\pi(x,z)) &= \sum_x \pi(x)(1-\pi(x)) = K - \sum_x \pi(x)^2.
\end{align*}
Using this fact above yields 
\begin{align*}
\Var(\htotipw)&\leq  A^2\exp(-2\modelmin/\beta)\bigg(K-\sum_x \pi(x)^2\bigg) \\
&= A^2\exp(-2\modelmin/\beta)\bigg(K-\frac{1}{A^2}\sum_x \exp(\frac{2\model(x)}{\beta})\bigg) \\
&=\exp(-2\modelmin/\beta)\bigg(A^2K-\sum_x \exp(\frac{2\model(x)}{\beta})\bigg) \\ 
&= \exp(-2\modelmin/\beta)\bigg(\frac{1}{K}\sum_{x,z}\exp(\frac{\model(x)+\model(z)}{\beta})-\sum_x \exp(\frac{2\model(x)}{\beta})\bigg), 
\end{align*}
which is as desired. 

\paragraph{DR Estimation.}
For the DR estimator -- assuming rewards and model predictions are bounded in [0,1] -- $\theta(x)=r(x)-\model(x)\in[-1,1]$, so 
\begin{align*}
\bigg(\frac{r(x)-\model(x)}{\pi(x)} - \frac{r(z)-\model(z)}{\pi(z)}\bigg)^2 &= A^2 \bigg(\frac{r(x)-\model(x)}{\exp(\model(x)/\beta)} - \frac{r(z)-\model(z)}{\exp(\model(z)/\beta)}\bigg)^2 \\ 
    &= A^2\bigg(\frac{(r(x)-\model(x))^2}{\exp(2\model(x)/\beta)} + \frac{(r(z)-\model(z))^2}{\exp(2\model(z)/\beta)} 
    -2\frac{(r(x)-\model(x))(r(z)-\model(z))}{\exp(\frac{\model(x)-\model(z)}{\beta})}\bigg) \\ 
    &\leq 2A^2\bigg(\max_x \bigg\{\frac{(r(x)-\model(x))^2} {\exp(\frac{2\model(x)}{\beta})}\bigg\} + \bigg\vert \frac{(r(x)-\model(x))(r(z)-\model(z))}{\exp(2\frac{\model(x)+\model(z)}{\beta})}\bigg\vert\bigg) \\ 
    &\leq 2A^2\bigg(\max_x \bigg\{\exp(-2\model(x)/\beta)\bigg\} + \bigg\vert \exp(-2\frac{\model(x)+\model(z)}{\beta})\bigg\vert\bigg) \\ 
    &\leq 2A^2\bigg(\max_x \bigg\{\exp(-2\model(x)/\beta)\bigg\} + \max_x \exp(-2\model(x)/\beta)\bigg) \\
    &= 4A^2\exp(-2\modelmin/\beta).
\end{align*}
From here, the proof is identical to the IPW case but with an extra factor of 2. 

\subsection{Proof of Proposition~\ref{prop:var_bound_kl}}
Almost identical to the Entropy sampling  case, with extra multiplicative factors of $\model(x)/\modelmin$. 

\subsection{Proof of Proposition~\ref{prop:hoeffding_ipw}}
In a fixed period $T$, we consider running the sampling procedure $m$ times. We condition on the rewards and the model up until this point, so the only randomness stems from the selection, i.e., randomness over $S=S_T$. We will consider Entropy sampling; KL sampling is similar. 

We are concerned with concentration of the random variables $\htotipw = \htotipw(T)$, which are themselves sums of the random variables $\ind_{x\in S}$. Since $r(x)\in[0,1]$, we have $\htotipw\geq 0$. Also, 
\begin{align*}
    \htotipw = \sum_x \frac{r(x)}{\hpi(x)}\ind_{x\in S} \leq \sum_x \frac{1}{\hpi(x)}\ind_{x\in S}, 
\end{align*}
and 
\begin{align*}
    \hpi(x) = \frac{K\exp(\model(x)/\beta)}{\sum_z \exp(\model(z)/\beta)}\geq \frac{K\exp(\modelmin/\beta)}{\sum_z \exp(\model(z)/\beta)},
\end{align*}
where $\modelmin = \min_x\model(x)$. Hence, 
\begin{align*}
    \htotipw \leq \frac{\Gamma(\model)}{K} \sum_x\ind_{x\in S} = \Gamma(\model),
\end{align*}
where $\Gamma(\model) = \sum_z \exp(\model(z)/\beta) / \exp(\modelmin/\beta)$. This bound enables us to employ Hoeffding's concentration inequality on the $m$ iid random variables $\htotipw^1,\dots,\htotipw^m$, i.e., the result of running the sampling procedure $m$ times independently. Recall that for bounded random variables $Z_1,\dots,Z_m$ with $Z_i\in[0,B]$ almost surely, Hoeffding gives 
\[\Pr\bigg(\frac{1}{m}\bigg|\sum_i (Z_i - \E[Z_i])\bigg| \geq a\bigg) \leq 2\exp(-\frac{2ma^2}{B^2}).\]
Employing the bound above on $\htotipw$, and applying Hoeffding to the complement of the event gives 
\begin{align*}
    \Pr\bigg(\frac{1}{m}\bigg|\sum_i \htotipw^i - \E[\htotipw^i]\bigg| < a\bigg) \geq 1 - 2\exp(-\frac{2ma^2}{\Gamma^2(\model)}).
\end{align*}
Note that $\E[\htotipw^i] = \sum_x \frac{r(x)}{\hpi(x)}\E[\ind_{x\in S}] = \sum_x \frac{r(x)}{\hpi(x)}\pi(x)=\sum_x r(x)\lambda(x)$. Putting 
\[\delta = 2\exp(-\frac{2ma^2}{\Gamma^2(\model)}),\]
i.e., 
\[a^2 =  \frac{\Gamma^2(\model)}{2m}\log(\frac{2}{\delta}),\]
we obtain that with probability at least $1-\delta$, 
\begin{align*}
    \bigg|\frac{1}{m}\sum_i \htotipw^i - \sum_x r(x)\lambda(x)\bigg|< \frac{\Gamma(\model)}{\sqrt{2m}} \log^{1/2}\bigg(\frac{2}{\delta}\bigg).
\end{align*}
If $\lambda(x)=1$, then this completes the proof since $\sum_x r(x)=\tot$. If not, then we employ the asymptotic bound on $\lambda(x)$ developed by \citet{rosen2000inclusion} which, we recall, states that $\lambda(x)=1 + \varepsilon(x)$, where $\varepsilon(x)$ is an error term which tends to zero as $\log K/\sqrt{K}$. Hence,
\begin{equation}
\label{eq:rxlx}
  \sum_x r(x)\lambda(x) = \tot + \sum_xr(x)\epsilon(x) = \tot + O(\log K /\sqrt{K})\tot.  
\end{equation}
Let $E(K)$ be the term such that $\sum_xr(x)\lambda(x) = \tot + E(K)\tot$, and let $C$ be the constant being hidden in the $O$ such that  $|E(K)|\leq C\log K/\sqrt{K}$ for all sufficiently large $K$. Hoeffding's one-sided bound applied to Equation~\eqref{eq:rxlx} gives 
\begin{align*}
    \Pr\bigg(\frac{1}{m}\sum_i \htotipw^i - \tot + E(K)\tot \geq a \bigg)\leq \exp(-\frac{2ma^2}{\Gamma^2(\model)}). 
\end{align*}
Notice that $a-E(K)\tot \leq a + C\log K /\sqrt{K}\tot$ (for large enough $K$), hence 
\begin{align*}
    &\Pr\bigg(\frac{1}{m}\sum_i\htotipw^i - \tot \geq a + C\log K/\sqrt{K}\bigg) \\
    & \leq 
    \Pr\bigg(\frac{1}{m}\sum_i \htotipw^i - \tot \geq a - E(K)\tot\bigg) \\ 
    &\leq \exp(-\frac{2ma^2}{\Gamma^2(\model)}). 
\end{align*}
Likewise, the one-sided bound for $-a$ gives  
\begin{align*}
    \Pr\bigg(\frac{1}{m}\sum_i \htotipw^i - \tot + E(K)\tot \leq  -a \bigg)\leq \exp(-\frac{2ma^2}{\Gamma^2(\model)}). 
\end{align*}
Since $-a-C\log K/\sqrt{K}\tot \leq -a - E(K)\tot$, we have 
\begin{align*}
&\Pr\bigg(\frac{1}{m}\sum_i \htotipw^i - \tot \leq  -a - C\log K/\sqrt{K} \tot \bigg) \\
&\leq 
    \Pr\bigg(\frac{1}{m}\sum_i \htotipw^i - \tot  \leq  -a - E(K)\tot\bigg) \\ 
    &\leq \exp(-\frac{2ma^2}{\Gamma^2(\model)}). 
\end{align*}
Combining both inequalities furnishes the conclusion that, for large enough $K$, 
\begin{align*}
    \Pr\bigg(\bigg|\frac{1}{m}\sum_i\htotipw^i - \tot \bigg| \geq a + C\log K/\sqrt{K}\cdot \tot\bigg)\leq 2\exp(-\frac{2ma^2}{\Gamma^2(\model)}).
\end{align*}
From here, the result follows in the same way as above.

\section{Variance Calculations}
\label{app:variance}

For any $\theta:\X\to\R$ set
\[Q(\theta) = \sum_{x} \frac{\theta(x)}{\hpi(x)}\delta_S(x),\]
where $\delta_S(x)$ indicates whether $x$ is in the random set $S$. 

The variance with respect to $S$ is  
\begin{align*}
\Var_S(Q(\theta)) &= 
\sum_{x}\frac{\theta^2(x)}{\hpi(x)^2}\Var(\delta_S(x)) + \sum_{x}\sum_{z\neq x}\frac{\theta(x)\theta(z)}{\hpi(x)\hpi(z)}\Cov(\delta_S(x),\delta_S(z)) \\
&= \sum_{x}\frac{\theta^2(x)}{\hpi(x)}(1-\pi(x)) + \sum_{x}\sum_{z\neq x}\frac{\theta(x)\theta(z)}{\hpi(x) \hpi(z)}(\pi(x,z)-\pi(x)\pi(z)), 
\end{align*}
since $\Var(\delta_S(x))) = \E[\delta_S(x)] - \E[\delta_S(x)]^2 = \pi(x)(1-\pi(x))$ and similarly for the covariance. (Note that $\ind(x\in S) = \ind(x\in S)^2$.) Appendix~\ref{app:incl_probs} gives 
\[\sum_{z\neq x}(\pi(x,z)-\pi(x)\pi(z)) = (K-1)\pi(x) - \pi(x)(K-\pi(x)) = \pi(x)(\pi(x)-1).\]

Plugging this into the first sum of $\Var(Q(z))$ we have 
\begin{align*}
\Var(Q(\theta)) &= -\sum_{x}\frac{\theta^2(x)}{\hpi(x)\pi(x)}\sum_{z\neq x}(\pi(x,z)-\pi(x)\pi(z))+ \sum_{x}\sum_{z\neq x}\frac{\theta(x)\theta(z)}{\hpi(x) \hpi(z)}(\pi(x,z)-\pi(x)\pi(z))\\
&= \sum_{x}\sum_{z\neq x}\bigg(\frac{\theta(x)\theta(z)}{\hpi(x) \hpi(z)}-\frac{\theta^2(x)}{\pi(x)\hpi(x)}\bigg)(\pi(x,z)-\pi(x)\pi(z)). 
\end{align*}
If we assume that $\hpi(x)=\pi(x)$, then we can write 
\begin{align*}
\Var(Q(\theta)) &= \sum_{x}\sum_{z\neq x}\bigg(\frac{\theta(x)\theta(z)}{\pi(x) \pi(z)}-\frac{\theta^2(x)}{\pi(x)^2}\bigg)(\pi(x,z)-\pi(x)\pi(z)) \\ 
&= \frac{1}{2}\sum_{x}\sum_{z\neq x}\bigg(\frac{\theta(x)}{\pi(x)}-\frac{\theta(z)}{\pi(z)}\bigg)^2(\pi(x)\pi(z)-\pi(x,z)).
\end{align*}
In the last line we completed the square and divided by two to account for the extra terms being introduced. Substituting the relevant $\theta$ for either estimator gives the desired result.

\section{Adaptive Bin Sampling}
\label{app:abs}

At each timestep, ABS first greedily selects the $Z<K$ observations with the highest predictions. For the remaining $K-Z$ selections, ABS parameterizes the model predictions using either a logistic or an exponential function. In our notation, observation $x$ is assigned the value 
\[v_\alpha(x) = ( 1 + \exp(-\alpha(\model(x) - \kappa)),\] 
(for logistic) or 
\[v_\alpha(x)=\exp(\alpha x),\] (for exponential) 
where $\alpha\in\R$ is a hyperparameter and $\kappa$ represents the value of the $K$-th largest prediction. 
ABS normalizes the predictions such that  $\model(x)\in[-5,5]$ for logistic smoothing,  and $\model(x)\in[0,1]$ for exponential smoothing. 
The resulting values are clustered into $H$ non-overlapping strata ($H$ is chosen beforehand), ensuring that there are at least $K$ observations per cluster. Formally, ABS solves (or approximates) the constrained optimization problem 
\begin{equation*}
    \min_{C_1,\dots,C_H} \quad \sum_{h} \sum_{v_\alpha(x)\in C_h} \norm{\rho(x) - A_h}^2, \quad
    \textnormal{s.t.} \quad |S_h|\geq K-Z,
\end{equation*}
using a modified form of $K$-means, where 
\begin{equation*}
    A_h = \frac{1}{|C_h|}\sum_{x\in C_h}v_a(x).
\end{equation*}
Cluster $C$ is then assigned a sampling probability $\pi(C)$ proportional to $A_h$, with the caveat that $\pi(C)\geq \tau$ where $\tau\geq 0$ is a parameter called the ``trim factor''. The trim factor ensures that no strata has too low of a sample probability. 
Observations are selected by drawing from the distribution $(\pi(C_1),\dots,\pi(C_H))$ a total of $K-Z$ times \emph{with replacement}, and then selecting observations within each strata uniformly at random \emph{without replacement}. This two-stage sampling strategy enables observations to be sampled using precise inclusion probabilities (i.e., the Pareto sampling approximation is not needed). However, within clusters, each observation has the same probability of selection. Moreover, each cluster must contain at least $K-Z$ observations. 

If $x\in C_h$, then 
\[\pi(x) = \frac{K-Z}{|C_h|}\pi(C_h).\]
Let $D$ be the set of $Z$ observations for chosen. For $x\in D$ we have $\pi(x)=1$. Thus, the IPW estimator for ABS is 
\begin{equation*}
    \htotipw(t) = \sum_{h}\sum_{x\in S_h} \frac{\re(x)}{\pi(x)} + \sum_{x\in D}\re(x).
\end{equation*}
Throughout our experiments, we use $H=10$ and $Z=0.1K$. We use a trim of 0 for most experiments, but we explore its effects in Appendix~\ref{app:results}.

\section{Why can't we minimize variance directly?}
\label{app:intractable}
Here we consider the objective function formed via a linear combination of variance and expected reward:   
\[\Phi(\pi) = \E[\reward(\pi,\model)] - \beta \Var(\htot),\]
where, for the IPW estimator for example, 
\[\Var(\htotipw) = \frac{1}{2}\sum_{x,z}\bigg(\frac{r(x)}{\pi(x)} - \frac{r(z)}{\pi(z)}\bigg)^2(\pi(x)\pi(z)-\pi(x,z)).\]
As mentioned above, calculating (let alone optimizing) the joint inclusion probabilities is difficult. Often, they cannot be computed in closed-form. Even when they can be, they are unwieldy to compute. The joint inclusion probabilities for conditional Poisson Sampling, for instance, are 
\[\pi_{i,j} = \frac{\sum_{s \in A(i,j)}\prod_{i\in s}p_i\prod_{i\notin s}(1-p_i)}{\sum_{s \in A}\prod_{i\in s}p_i\prod_{i\notin s}(1-p_i)}.\]
These can be computed by a recursive procedure~\citep{aires1999algorithms} and avoid exponential time, but optimizing over the values $p_1,\dots,p_n$ would be challenging. 

Instead of considering the joint inclusion probabilities, two other options are to (i) sample with replacement, or (ii) approximate the joint inclusion probabilities with a simpler expression. 
(i) is undesirable from a practical standpoint, especially for smaller sample sizes. Companies and public agencies often have fixed budget constraints which are given a priori \cite{henderson2021beyond}, and sampling with replacement risks under-utilizing the budget. One option is to simply repeat the sampling procedure until the budget is fully used. For example, if $K_1<K$ observations are sampled during the first iteration, then repeat the process with a new budget of $K-K_1$, and so on. However, the problem of calculating the inclusion probabilities rears its head here once again, as the true inclusion probabilities must now take into account this iterative process. The final difficulty with this direction is that it does not yield a well-behaved objective function. The (estimated) variance of sampling with replacement is
\[\sum_{x}\model(x)(\pi(x)^{-1}-1),\]
so the objective function with the Lagrangian constraint becomes 
\[\min_\pi \bigg\{\sum_{x}\model(x)(\pi(x)^{-1}-1)- \sum_x \pi(x)\model(x) + \lambda(\sum_x\pi(x)-1)\bigg\}.\]
Taking the derivative w.r.t $\pi(x)$ and setting to 0 gives 
\[\pi(x) = \bigg(\frac{1}{1-\lambda/\model(x)}\bigg)^{1/2}.\]
Thus, it's impossible to solve for lambda explicitly by using the constraint $\sum_x\pi(x)=1$. Finally, we note that even if one could find a tractable optimization problem, sampling according to (approximations to the) prescribed joint inclusion probabilities is non-trivial and has been the subject of much attention in the survey sampling literature. See \citet{tille2006sampling} for a survey. 

Regarding (ii), the most well-known joint inclusion probability approximation comes from \citet{hajek1964asymptotic}, who proposed 
\begin{equation*}
    \tilde{\pi}(x,z) = \pi(x)\pi(z)(1 - (1-\pi(x))(1-\pi(z))L),
\end{equation*}
where 
\[L = \bigg(\sum_x \pi(x)(1-\pi(x))\bigg)^{-1}.\]
H\'ajek showed that this approximation obeys 
\begin{equation*}
    \pi(x,z) = \pi(x)\pi(z)\bigg(1 - L(1-\pi(x))(1-\pi(z))(1 + o(1))\bigg),
\end{equation*}
for rejective sampling procedure. \citet{berger1998rate} generalized the approximation error, demonstrating that for any sampling design, the relative error is o(1) plus the KL divergence between the distribution of the rejective sampling design and the given design. 

While $\tilde{\pi}(x,z)$ is a closed-form approximation of $\pi(x,z)$ involving only first-order inclusion probabilities, each term involves a sum over all such probabilities. Plugging this approximation into the objective function above does not, to put it mildly, yield a friendly optimization problem.





\section{Hyperparameter Tuning}
\label{app:tuning}

For hyperparameter tuning, we run a randomized grid search using five-fold CV for 50 iterations. The grid searches over number of estimators (from 100 to 2000, 12 evenly spaced options), the maximum depth of trees (from 10 to 110, by 10, and no max-depth), the minimum number of samples required to split an internal node (2, 5, and 10), the minimum number of samples required to be at a leaf node (1,2, and 4), and whether bootstrap samples are used when building trees (True, False).

The selected set of parameters for each dataset were: 

\paragraph{CPS.} 790 estimators, 10 minimum splitting samples, 4 minimum samples to be a leaf node, a max depth of 80 and bootstrapping. 

\paragraph{ACS.} 445 estimators, 10 minimum splitting samples, 4 minimum samples to be a leaf node, max depth of 20, and bootstrapping. 

\paragraph{Voting.} 272 estimators, 2 minimum splitting samples, 4 minimum samples to be a leaf node, max depth of 10, and bootstrapping. 

\paragraph{AllState.} 790 estimators, 2 minimum splitting samples, 4 minimum samples to be a leaf node, max depth of 110, and bootstrapping.

\section{Extended Results}
\label{app:results}

\begin{figure}
    \centering
    \subcaptionbox{ACS, Budget 1000}{\includegraphics[scale=0.4]{figures/reward_variance_PUMS_2014-18_ests__ipw_,_dr_,_model__period_4_budget_1000.png}}
    \subcaptionbox{AllState, Budget 1000}{\includegraphics[scale=0.4]{figures/reward_variance_Allstate_ests__ipw_,_dr_,_model__period_5_budget_1000.png}}
    \subcaptionbox{CPS, Budget 500}{\includegraphics[scale=0.4]{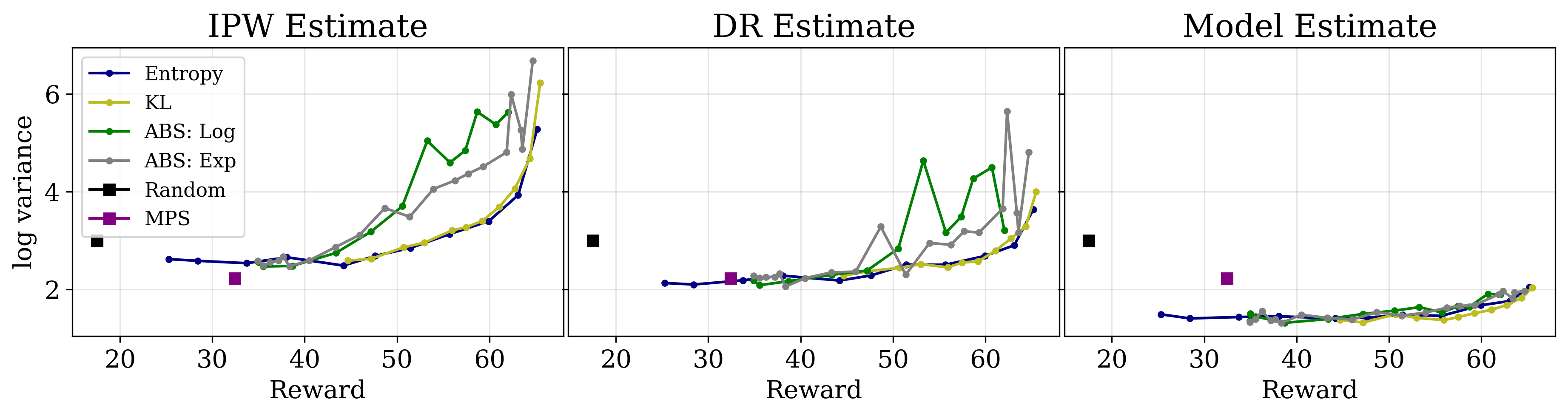}}
    \subcaptionbox{Voting, Budget 200}{\includegraphics[scale=0.4]{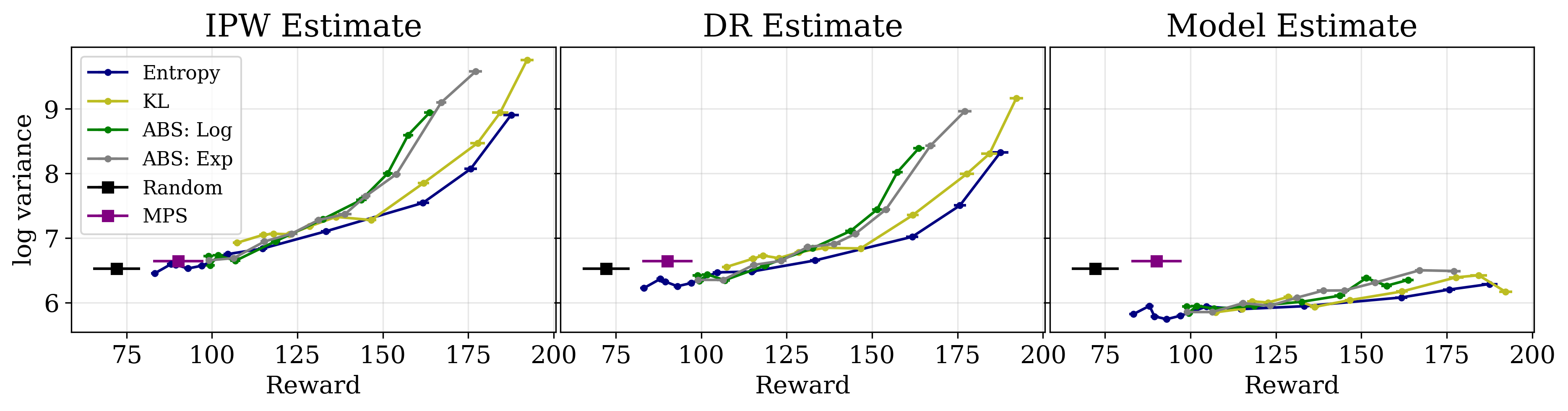}}
    \caption{Reward-variance trade off curves across all four datasets, plotted with 95\% confidence intervals on reward.}
    \label{fig:reward_variance_all_datasets}
\end{figure}

\paragraph{Improvement of Entropy and KL over ABS is consistent across all datasets.}
While the extent of the improvement varies somewhat across datasets, Entropy and KL consistently outperform ABS (with both logistic and exponential smoothing) for the IPW and DR estimators (Figure~\ref{fig:reward_variance_all_datasets}). When there is high variance among the model-based estimates (ACS data), Entropy and KL sampling continue to outperform ABS. Of course, when there is low variance among the model-based estimators (the remaining three datasets), the distinction between the four methods mostly vanishes. We discuss the variance of the model-based estimate next.

\begin{figure}
    \centering
    \subcaptionbox{Period 1}{\includegraphics[scale=0.4]{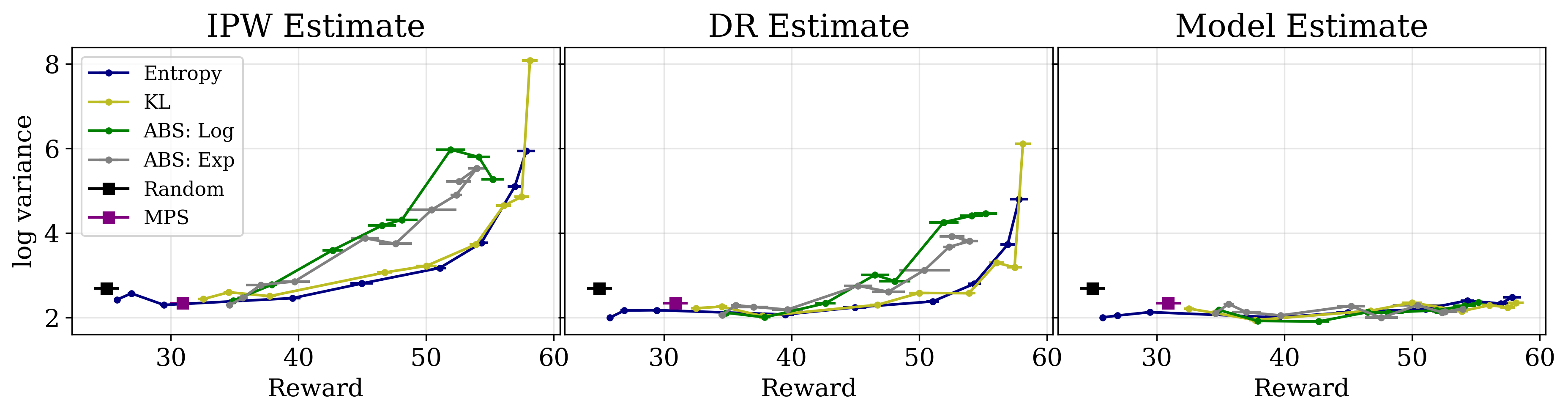}}    \subcaptionbox{Period 2}{\includegraphics[scale=0.4]{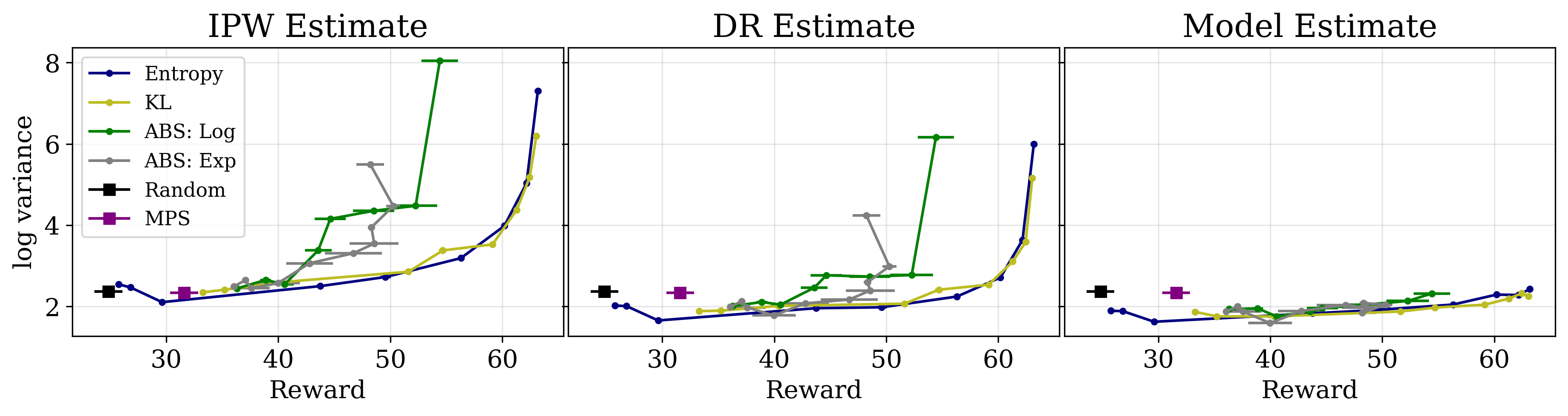}}
    \subcaptionbox{Period 3}{\includegraphics[scale=0.4]{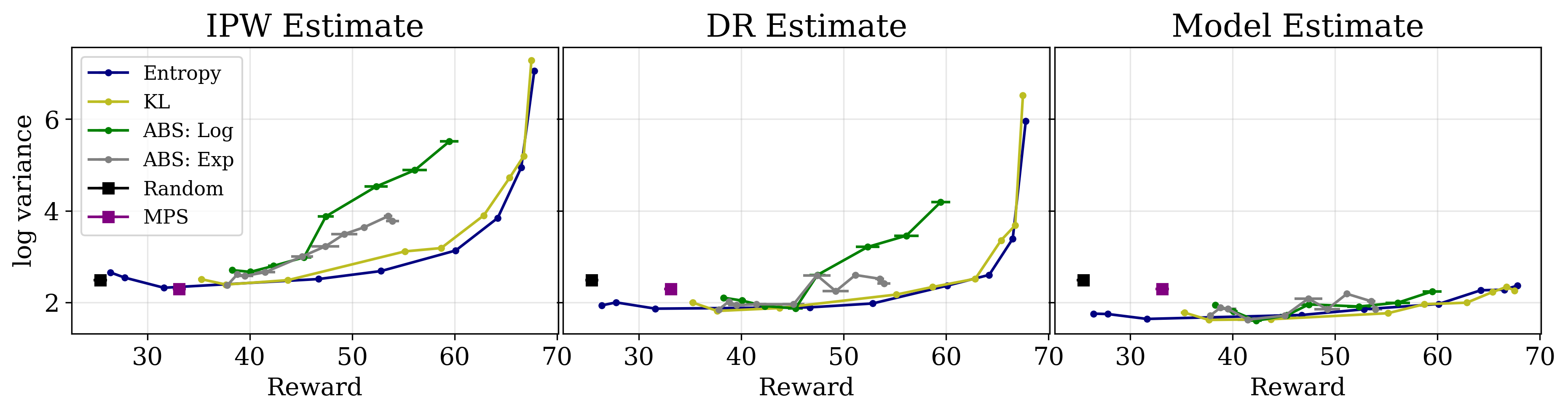}}
    \subcaptionbox{Period 4}{\includegraphics[scale=0.4]{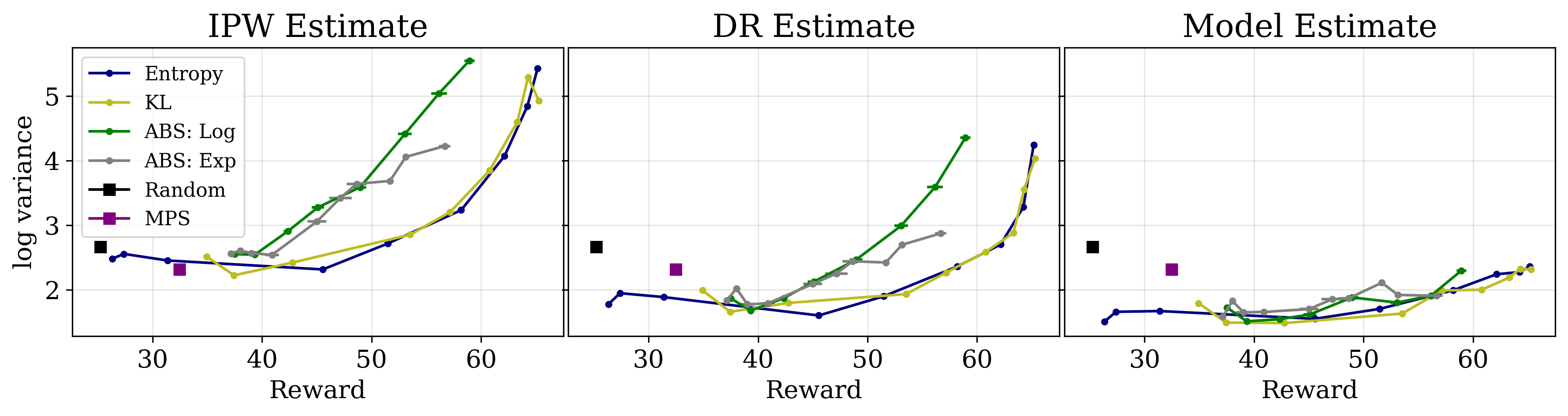}}
    \subcaptionbox{Period 5}{\includegraphics[scale=0.4]{figures/reward_variance_Allstate_ests__ipw_,_dr_,_model__period_5_budget_1000.png}}
    \caption{Reward Variance trade offs for periods one to five on AllState data with a budget of 1000, plotted with 95\% confidence intervals on reward.}
    \label{fig:allstate_periods}
\end{figure}

\paragraph{Entropy and KL Sampling are more consistent across timesteps than ABS.}
Figure~\ref{fig:allstate_periods} demonstrates the reward variance trade off of all the estimators across periods one to give for the AllState dataset. Period zero is not pictured, since observations were selected at random. While Entropy and KL sampling continue to outperform ABS, we see that the behavior of ABS with both logistic and exponential smoothing is more erratic than that of Entropy and KL. The latter demonstrate a relatively smooth reward-variance curve each timestep, while the former change the shape of their curves each timestep.

\begin{figure}
    \centering
    \includegraphics[scale=0.3]{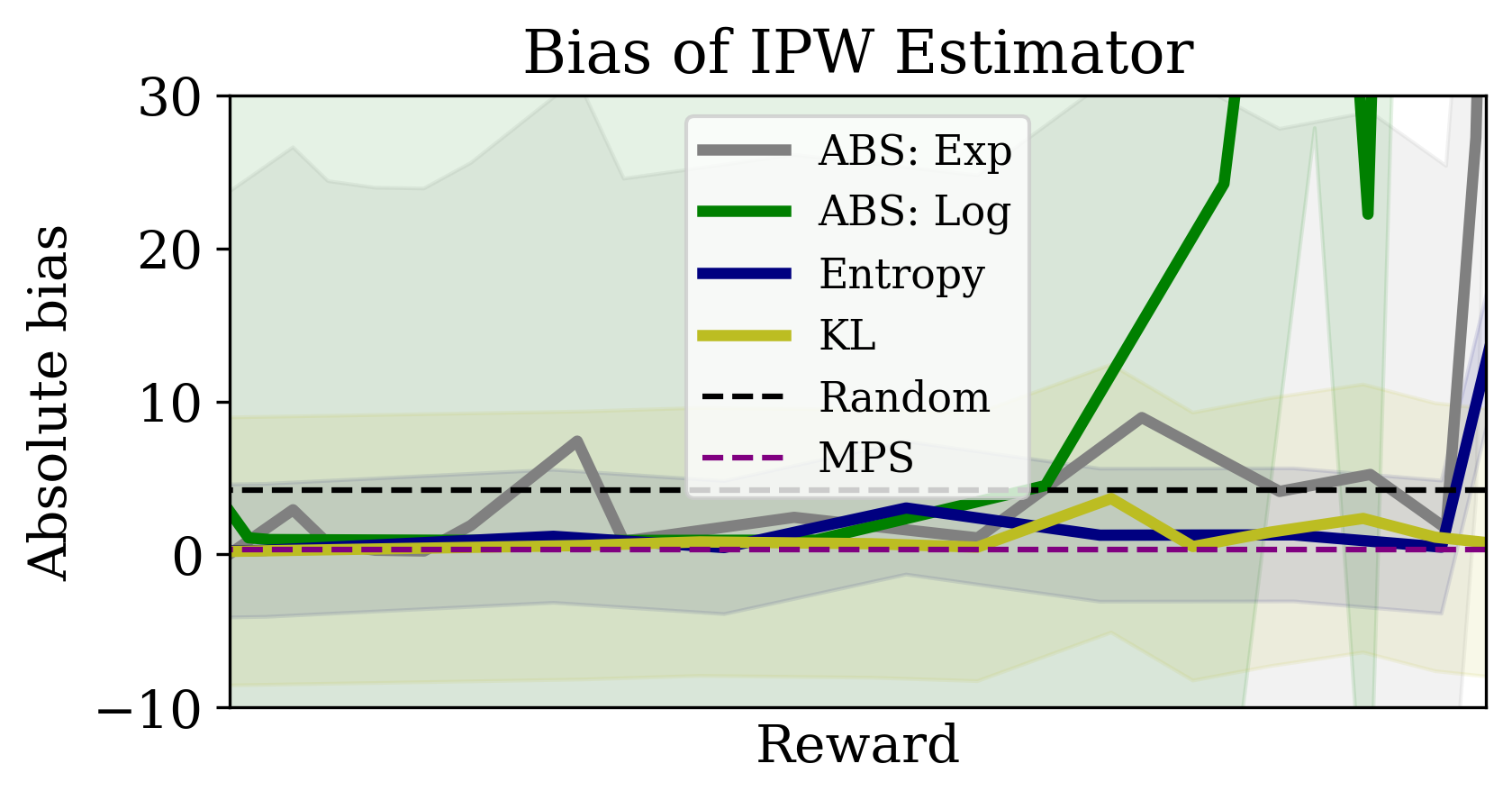}
    \includegraphics[scale=0.3]{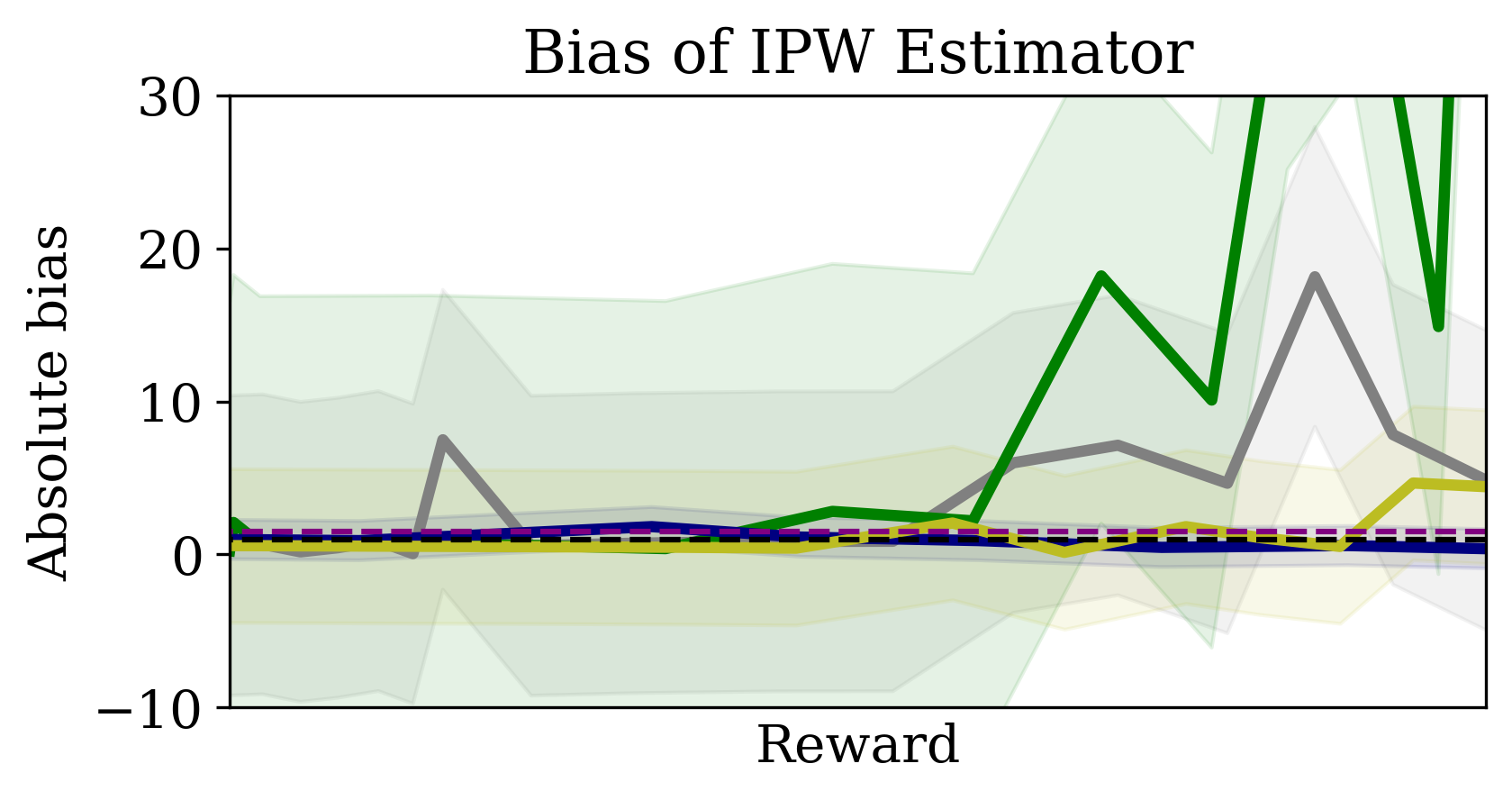}
    \includegraphics[scale=0.3]{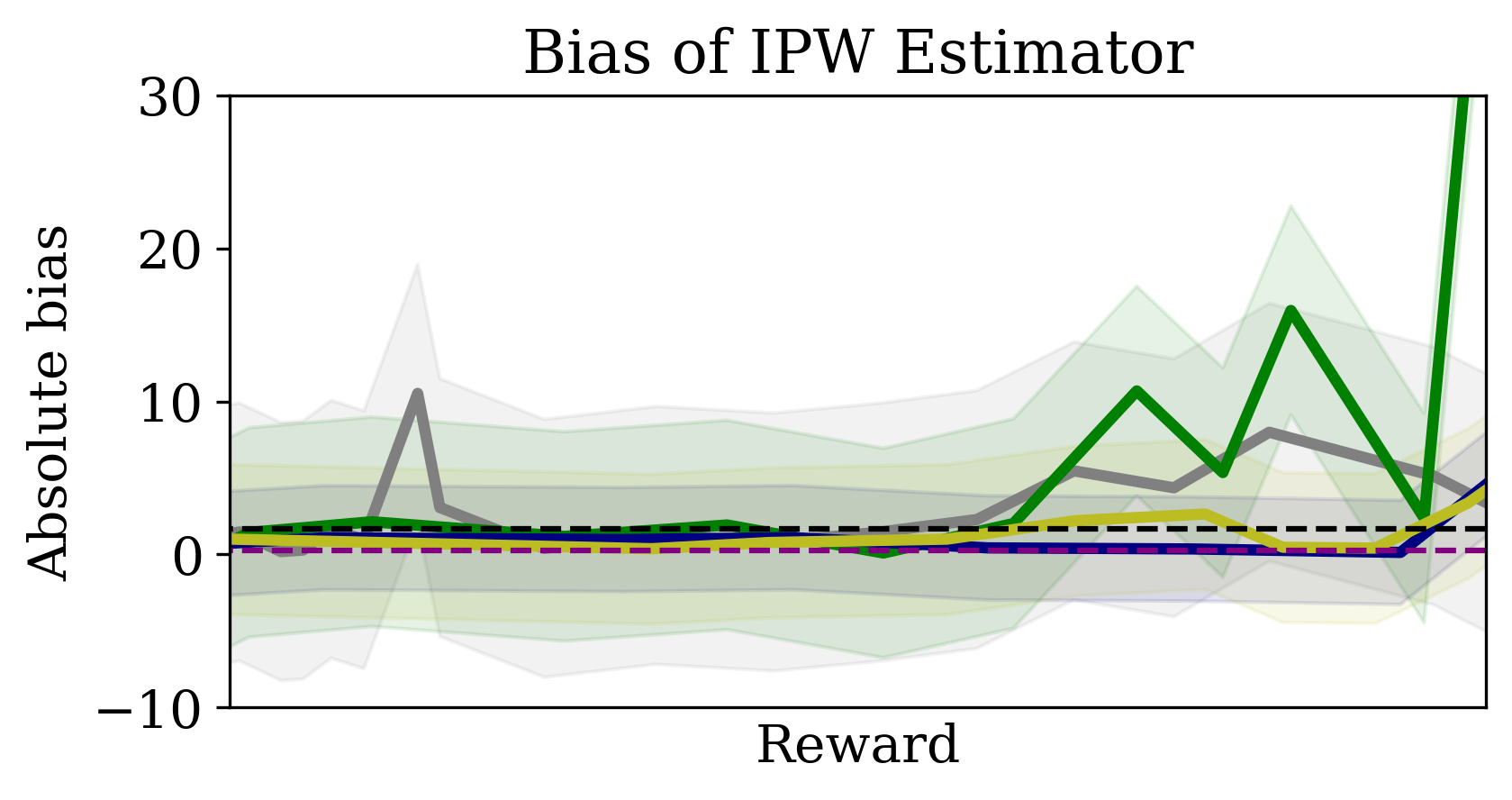}
    \includegraphics[scale=0.3]{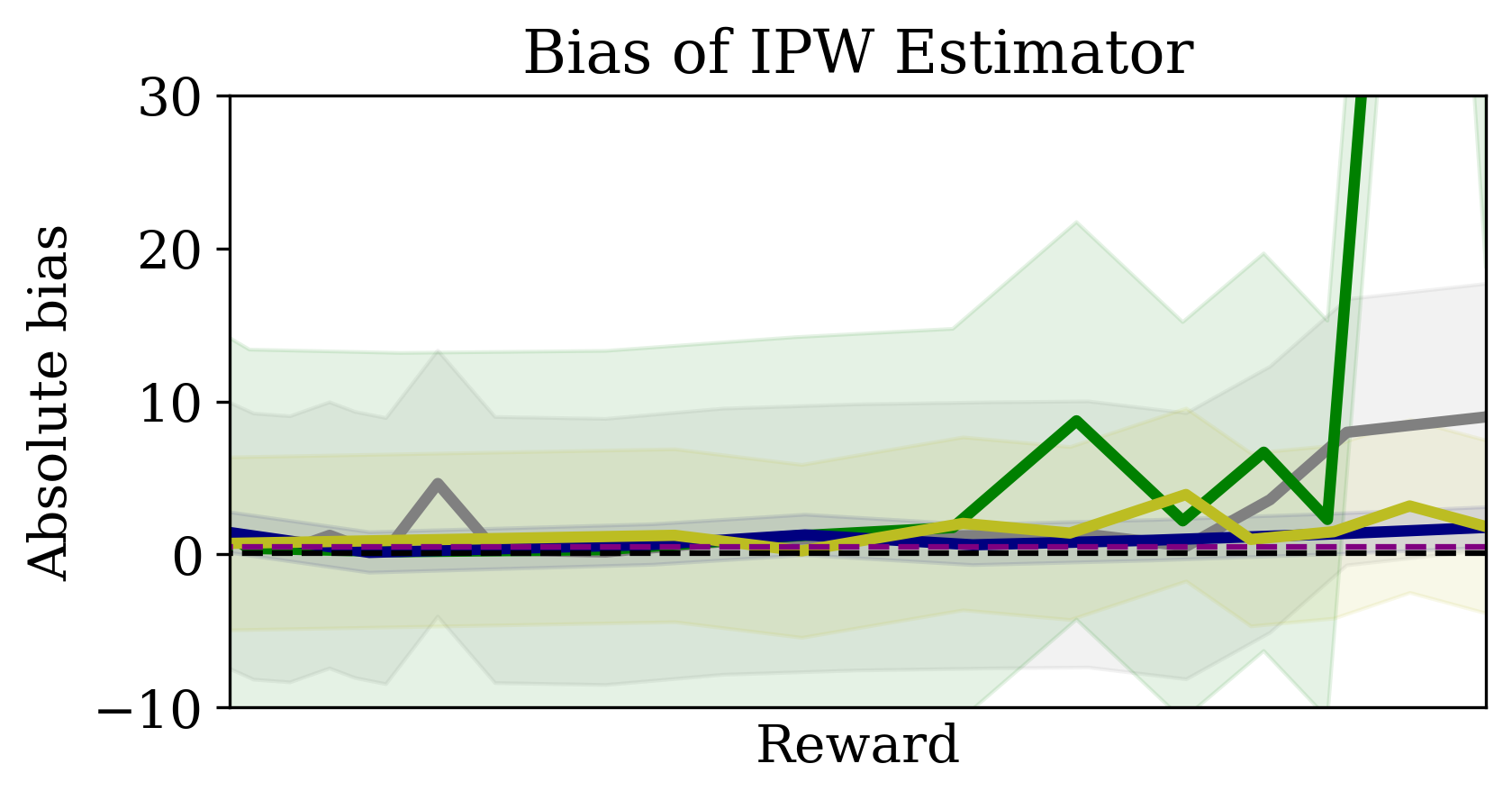}
    \includegraphics[scale=0.3]{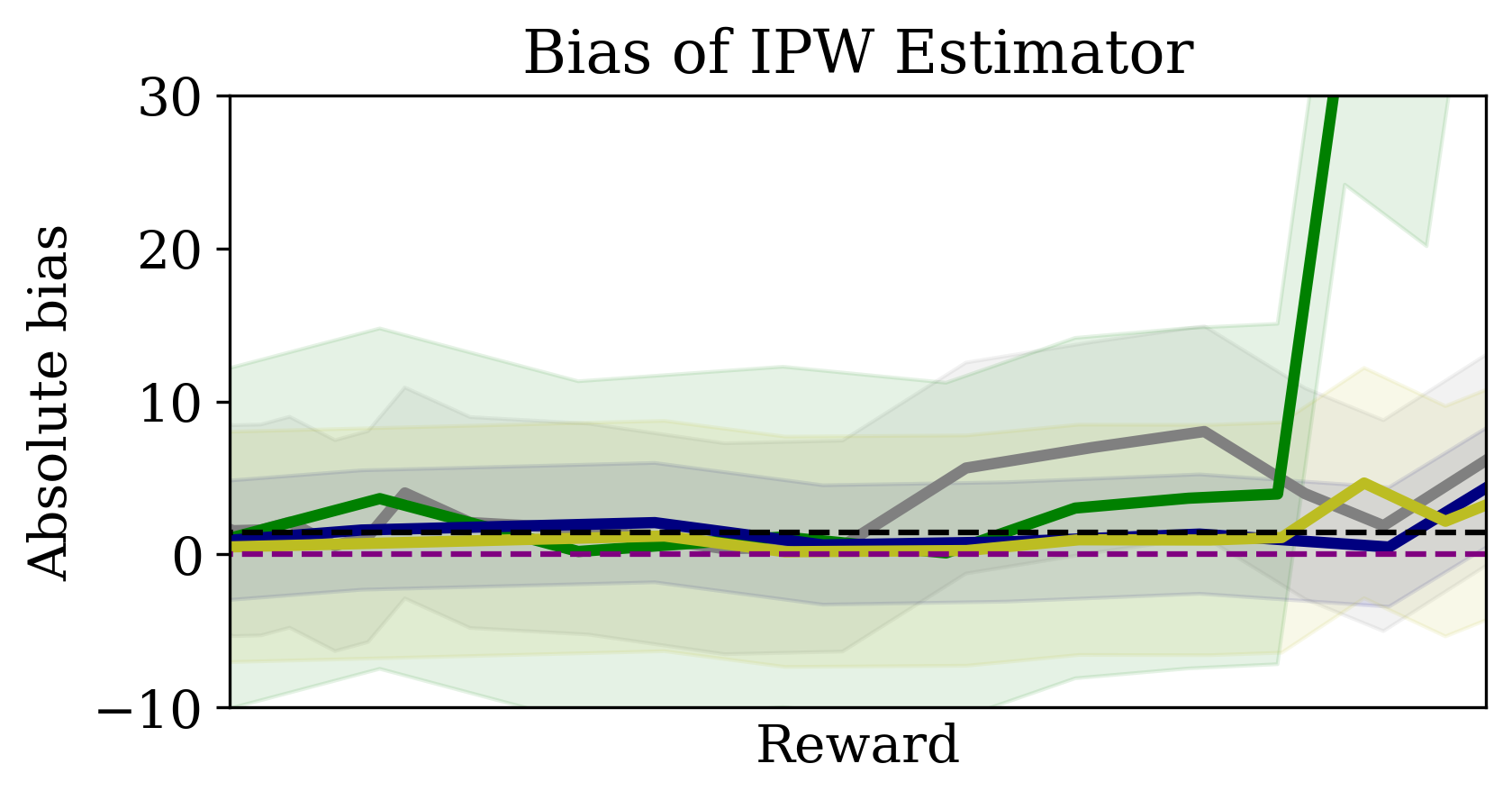}
    \includegraphics[scale=0.3]{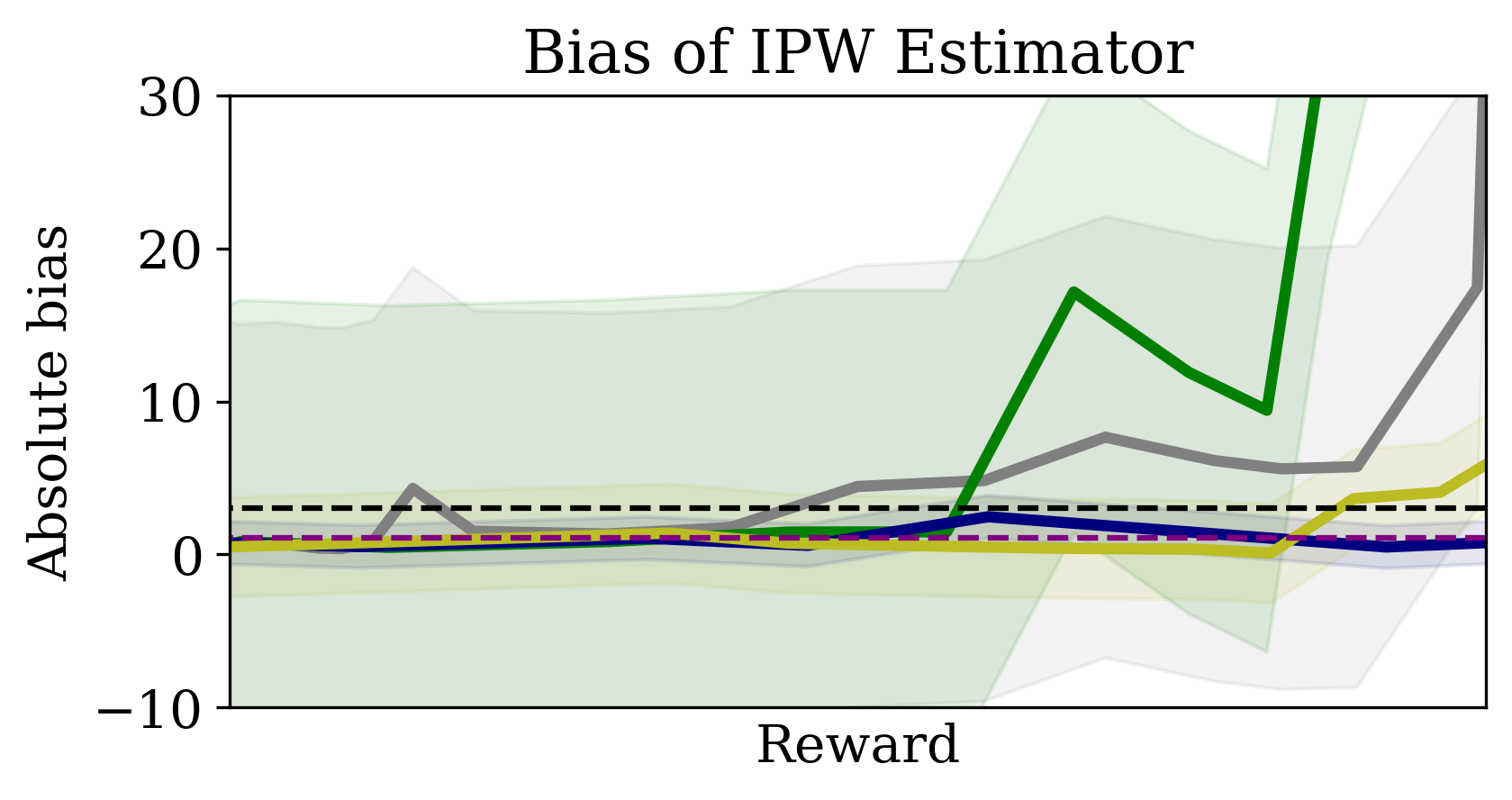}
    \caption{Bias of IPW estimator across sampling strategies on CPS data with a budget of 500. The bands around each line represent 95\% confidence intervals. }
    \label{fig:ipw_bias_timesteps}
\end{figure}

\paragraph{Bias is consistent across timesteps.}
Figure~\ref{fig:ipw_bias_timesteps} extends Figure~\ref{fig:bias_ipw} and demonstrates that the absolute bias of the IPW estimator for each sampling strategy is consistent across timesteps. It shows the bias on the CPS dataset from period two to period seven. We see that while all strategies have nearly zero bias in lower reward regions, both ABS strategies are unable to maintain this low bias as reward begins to increase. 

\begin{figure}
    \centering
    \subcaptionbox{Budget 500}{\includegraphics[scale=0.4]{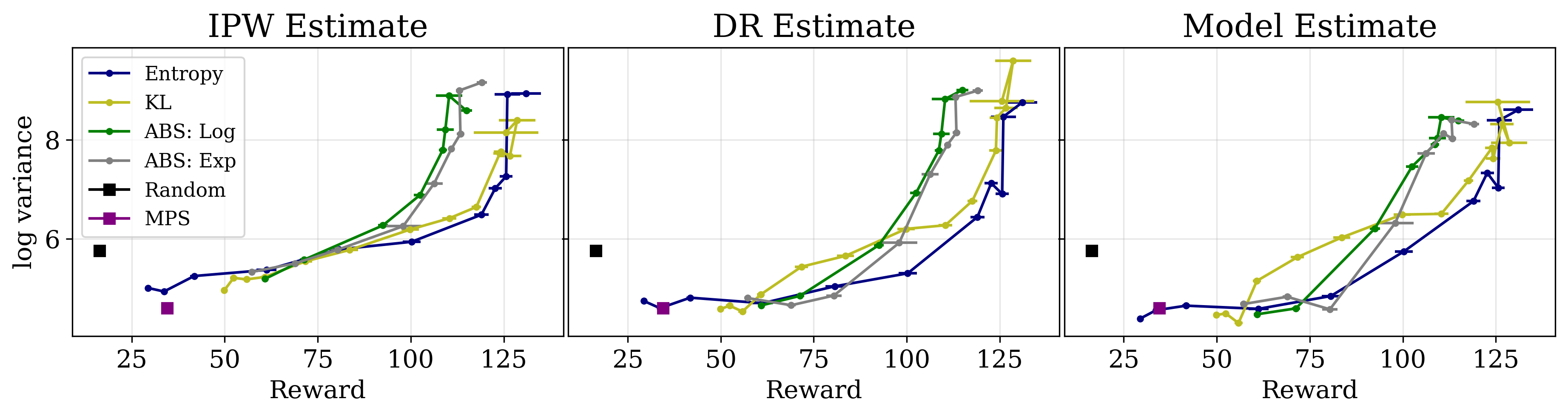}}
    \subcaptionbox{Budget 2000}{\includegraphics[scale=0.4]{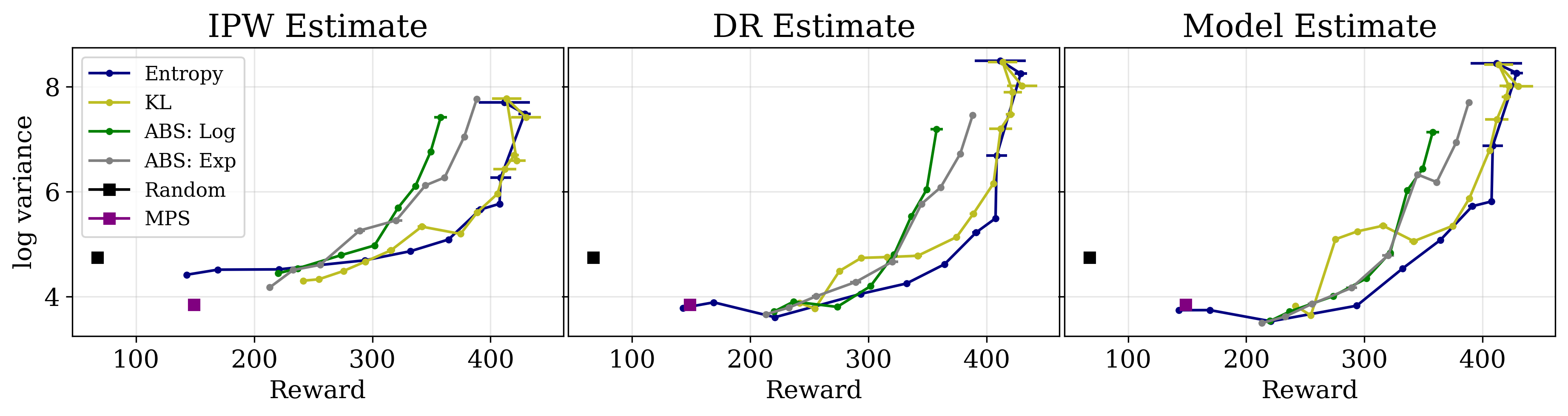}}
    \subcaptionbox{Budget 4000}{\includegraphics[scale=0.4]{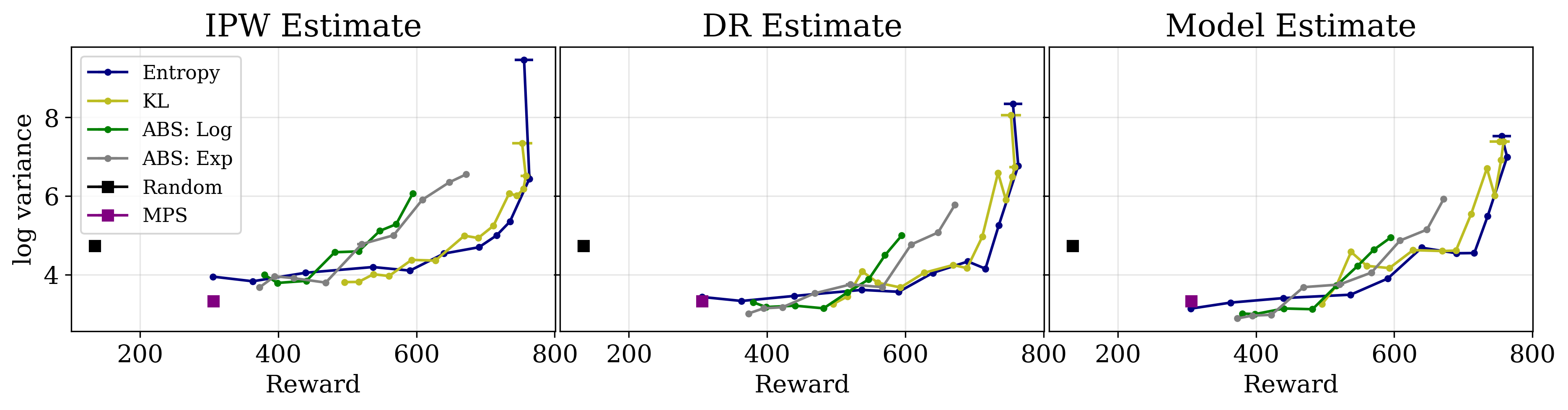}}
    \caption{The effect of various budgets on ACS data. Experiments were repeated 35 times and plotted with 95\% confidence intervals on reward.}
    \label{fig:budget}
\end{figure}

\paragraph{The Effect of Budget.} 
Figure~\ref{fig:budget} shows the reward-variance trade off for the ACS data on budget sizes of 500, 2000, and 4000. Experiments were repeated 35 times. Both budgets display the familiar reward-variance patterns, demonstrating that the effects are robust to budget sizes. We note that the variance of the model-based estimate begins to decrease with a budget of 4000. This corroborates the argument in Section~\ref{sec:results}, namely that a better fit model will exhibit lower variance (though, we emphasize, still have high bias). 

\begin{figure}
    \centering
    \begin{minipage}{0.49\textwidth}
    \subcaptionbox{\label{fig:trim_rv}}{\includegraphics[scale=0.4]{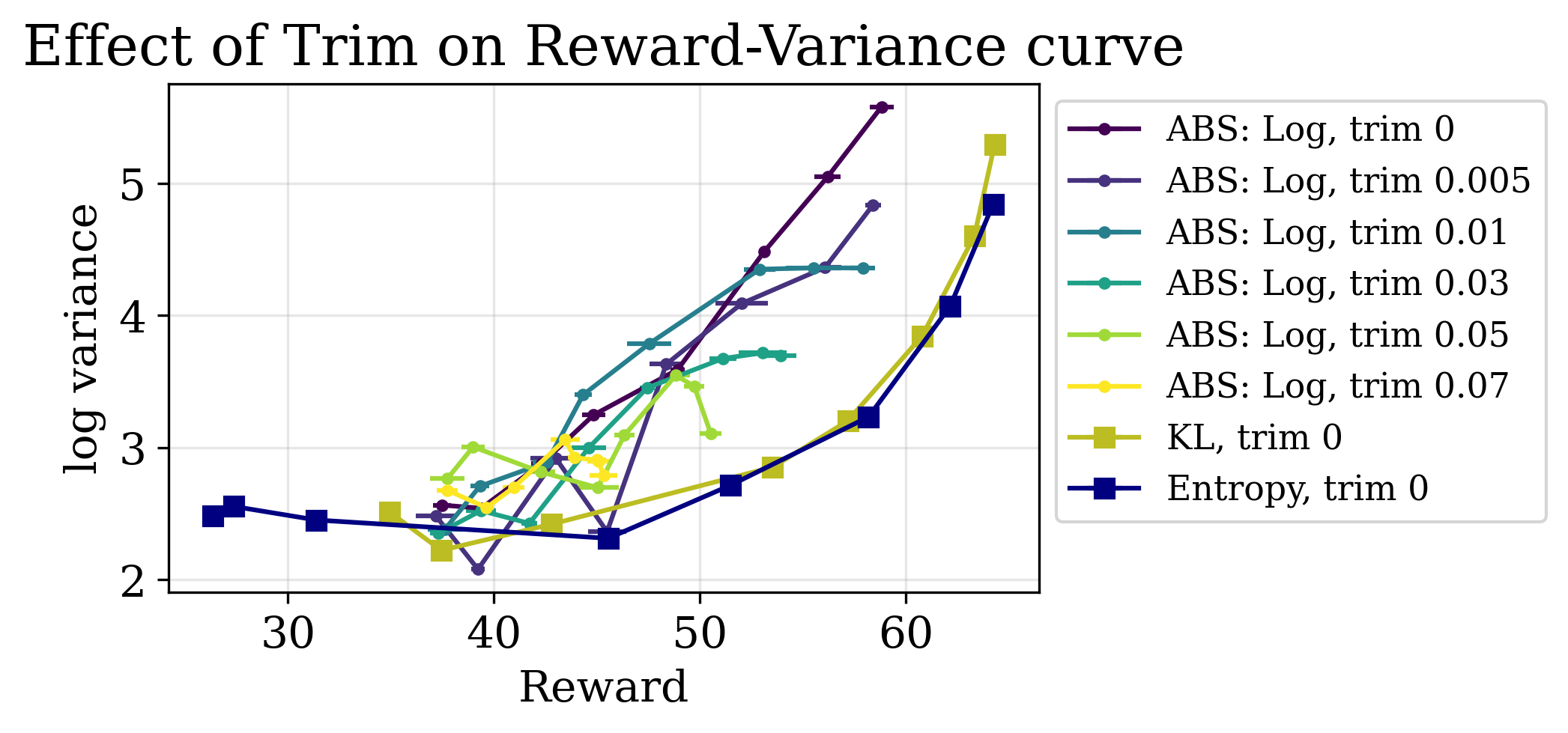}}
    \end{minipage}
    \begin{minipage}{0.49\textwidth}
    \subcaptionbox{\label{fig:trim_bias}}{\includegraphics[scale=0.4]{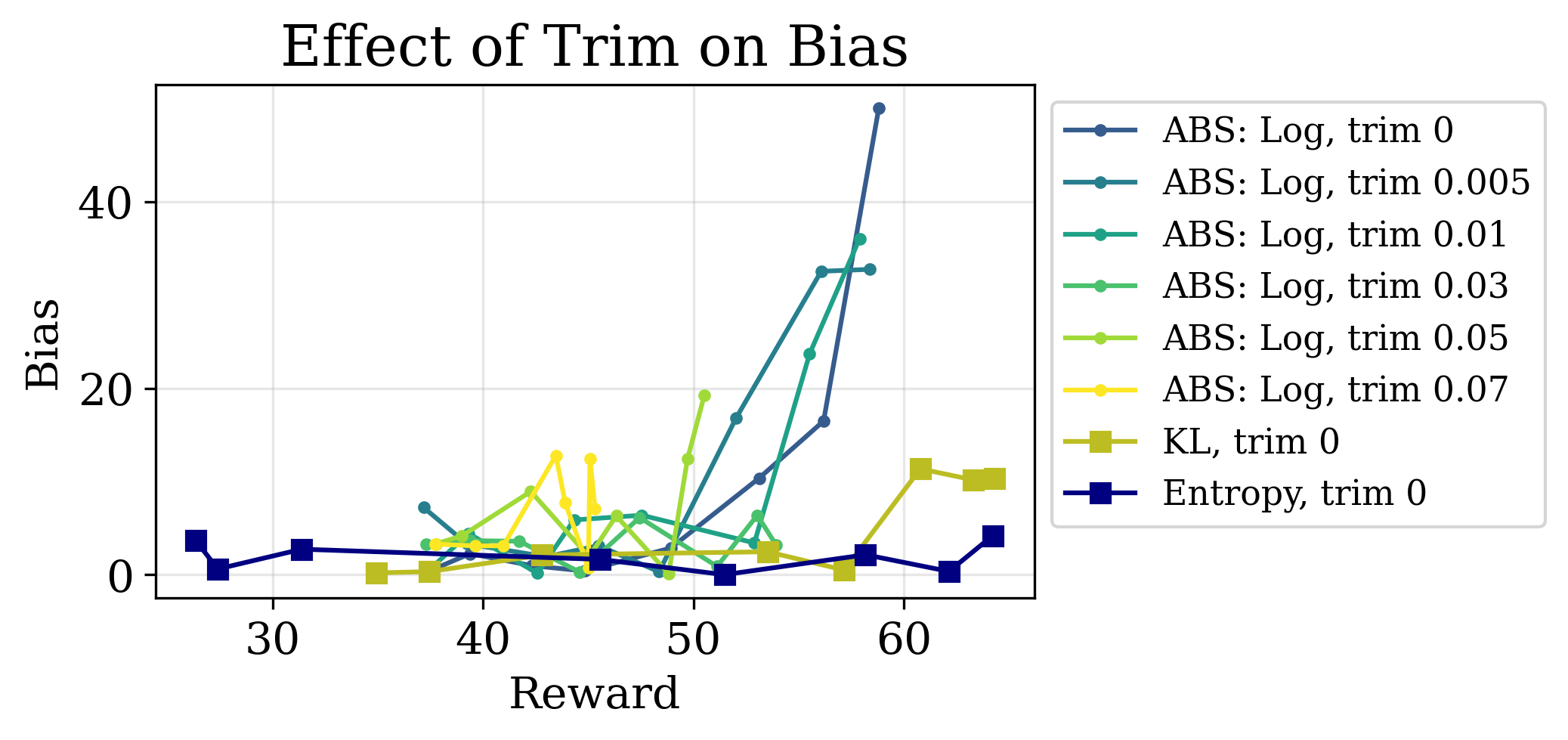}}
    \end{minipage}
    \caption{The effect of trim on reward, variance, and bias of ABS with logistic smoothing on the AllState budget with a budget of 1000. We compare the performance of ABS with trim to ABS with no trim and KL and Entropy Sampling.}
    \label{fig:trim_allstate}
\end{figure}

\paragraph{The Effect of Trim.}
Figure~\ref{fig:trim_rv} illustrates the effect of trim on ABS with logistic smoothing on the AllState dataset. At the top end of the reward range, adding trim does lower the variance. However, the drawback is that the reward-variance curves are shortened: We cannot reach the same reward levels without adding an unacceptable level of bias to the estimate. As the amount of trim gets higher, the curves therefore span less and less of the reward range. Figure~\ref{fig:trim_bias} demonstrates the bias across trim levels. We remove those parameters for which bias was too great, which occurred at lower levels of reward for higher trim factors.  

\end{document}